\documentclass[10pt,twocolumn,letterpaper]{article}

\usepackage{cvpr}              
\usepackage{multirow}
\usepackage{booktabs}
\usepackage{amsmath}
\usepackage{xcolor}
\usepackage{makecell}
\usepackage{svg}
\definecolor{cvprblue}{rgb}{0.21,0.49,0.74}
\usepackage[pagebackref,breaklinks,colorlinks,allcolors=cvprblue]{hyperref}

\def\confName{CVPR}
\def\confYear{2026}

\title{Dynamic Adversarial Reinforcement Learning for Robust Multimodal Large Language Models}
\author{
Yicheng Bao\textsuperscript{1} \quad
Xuhong Wang\textsuperscript{2,\textdagger} \quad
Qiaosheng Zhang\textsuperscript{2} \quad
Chaochao Lu\textsuperscript{2} \quad
Xia Hu\textsuperscript{2} \quad
Xin Tan\textsuperscript{1,2,\textdagger} 
\\ \\
\textsuperscript{1}East China Normal University \quad
\textsuperscript{2}Shanghai AI Laboratory \\
\vspace{0.3em}
\href{https://yicbao.github.io/To-Deceive-is-to-Teach-Forging-Perceptual-Robustness-via-Adversarial-Reinforcement-Learning}{Project Page}
}

\makeatletter
\def\@confName{} 
\def\@confYear{} 
\makeatother
\begin{document}


\twocolumn[{%
\renewcommand\twocolumn[1][]{#1}%
\maketitle
\begin{center}
    \includegraphics[width=0.94\textwidth]{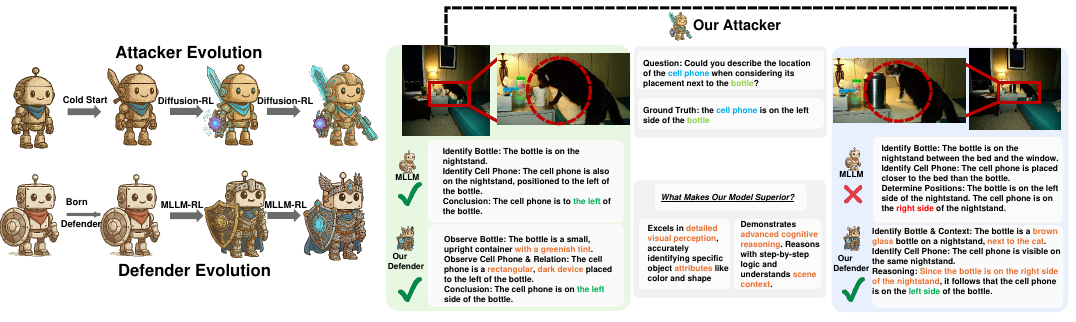}
    \captionof{figure}{An illustration of the \textbf{perceptual fragility} in existing MLLMs and the \textbf{robustness} fostered by our \textbf{co-evolutionary framework}. \textbf{Left (Co-evolution Concept):} A conceptual depiction of the \textbf{iterative competition} between an \textbf{Attacker} and a \textbf{Defender} model, where each model's capabilities are progressively enhanced. \textbf{Right (Practical Demonstration):} In a simple scene, both a standard MLLM and our Defender can correctly identify the spatial relation, but our model demonstrates superior, detailed perception. When a \textbf{contextual distractor} is introduced, the \textbf{standard MLLM} is \textbf{perceptually misled} by the distractor, causing its subsequent \textbf{reasoning to fail}. In contrast, our \textbf{Defender}, which shares the same base MLLM architecture, grounds its reasoning in a \textbf{robust perceptual understanding} of the scene, a capability directly fostered by our \textbf{adversarial training process}.}
    \label{fig:intro}
\end{center}
}]
\let\thefootnote\relax\footnotetext{\textsuperscript{\textdagger} Corresponding author.}

\begin{abstract}
Despite their impressive capabilities, Multimodal Large Language Models (MLLMs) exhibit perceptual fragility when confronted with visually complex scenes. This weakness stems from a reliance on finite training datasets, which are prohibitively expensive to scale and impose a ceiling on model robustness. We introduce \textbf{AOT-SFT}, a large-scale adversarial dataset for bootstrapping MLLM robustness. Building on this, we propose \textbf{AOT (Adversarial Opponent Training)}, a self-play framework that forges MLLM robustness by creating its own training data. Our method orchestrates a co-evolution between an image-editing Attacker and a Defender MLLM, where the Attacker generates a diverse and dynamic curriculum of image manipulations, forcing the Defender to adapt and improve. Extensive experiments demonstrate that AOT enhances the Defender's perceptual robustness and reduces hallucinations, establishing a scalable paradigm for training more reliable MLLMs.
\end{abstract}  

\section{Introduction}

Multimodal Large Language Model (MLLMs) have demonstrated remarkable advancements in comprehending and reasoning about the visual world \cite{ liu2023llava, bai_qwen25-vl_2025, zhu_internvl3_2025, hurst2024gpt, comanici2025gemini, guo2025deepseek}. By aligning powerful vision encoders with large language models (LLMs)\cite{yang2025qwen3, llama3modelcard}, these models have achieved state-of-the-art performance on a wide array of complex multimodal tasks.

Despite these impressive capabilities, the perceptual foundation of many leading MLLMs remains surprisingly fragile \cite{usama2025analysing, liu2025robustness, yang2024dynamic}. Their understanding of a visual scene can be easily compromised by minor modifications, a vulnerability that is particularly exacerbated in visually complex or cluttered scenes, where a high density of information provides more opportunities for contextual disruptions \cite{guo2024llava, zhang2024beyond}. This fragility is especially evident in tasks requiring fine-grained spatial perception \cite{chen2024spatialvlm, chen2025spatial}. As illustrated in \Cref{fig:intro}, a model that correctly identifies the relative positions of a cell phone and a bottle may fail completely when a canister is introduced nearby, causing it to mislocate the cell phone. This reveals a critical weakness: their grasp of visual relationships is not robust, undermining their ability to reason accurately about the visual world and raising fundamental questions about their reliability in real-world environments.

We argue that this fragility stems from a foundational issue in the prevailing training paradigm. The advancement of MLLMs is fundamentally tethered to large-scale, manually annotated datasets \cite{pattnayak2024survey, liang2024foundations}. This dependency creates a dual bottleneck: not only is the manual creation of such data prohibitively expensive and time-consuming, leading to a scarcity of examples for fine-grained perception \cite{wu2023multimodal}, but the resulting datasets are also inherently finite.
This finite nature imposes a \textbf{capability ceiling} on MLLMs. A model trained on a fixed dataset can only learn the patterns present within it, struggling to generalize to novel scenarios or withstand unforeseen variations \cite{villalobos2022will}. The limitation is especially acute for robustness enhancement, as finite adversarial datasets quickly become obsolete against continuously evolving models, failing to foster a truly resilient perceptual system \cite{pan2025beyond, stutz2020confidence, rice2020overfitting}. While dynamic evaluation protocols have been proposed \cite{yangDynamicMultimodalEvaluation2024b}, they are designed for assessment rather than training and lack a co-evolutionary mechanism to continually push the model's capabilities. The challenge, therefore, is not merely to acquire more data, but to fundamentally shift the training paradigm from a reliance on finite, pre-compiled corpora to one of autonomous and dynamic data generation. This shift would enable the creation of a continuous stream of relevant and challenging data without the bottleneck of massive human annotation efforts.

Inspired by the success of \textbf{self-play} in achieving superhuman performance through autonomous learning~\cite{wang2025vision, konyushkova2025vision, chen2023selfdebug, Pourcel2025SelfImprovingLM}, we propose a paradigm shift for enhancing MLLM perception. We introduce an iterative, generative adversarial framework where models improve by competing against each other, effectively creating their own training data. Our framework features a co-evolutionary dynamic between an \textbf{Attacker} ($M_{atk}$), an image editing model, and a \textbf{Defender} ($M_{def}$), the MLLM whose perception we aim to strengthen. This co-evolution forces the Defender to refine its perceptual acuity to overcome an ever-improving Attacker.

Our method uniquely operates by directly manipulating the \textit{image}, not the text. After an initial bootstrapping phase using a seed dataset, the framework enters a co-evolutionary cycle between an image-editing \textbf{Attacker} (\(M_{atk}\)) and the MLLM \textbf{Defender} (\(M_{def}\)). In each round, the Attacker is refined via policy optimization (Flow-GRPO \cite{liu2025flow}) to generate adversarial edits that can fool the current Defender. Crucially, these edits are guided by a reward function that jointly optimizes for adversarial efficacy and semantic integrity, the latter being enforced by a localized SSIM check. The updated Attacker then generates a curated set of challenging examples to fine-tune the Defender (via DAPO \cite{shao2024deepseekmath, yu2025dapo}). This iterative process, where the Attacker autonomously discovers a diverse spectrum of attack strategies from object removal to subtle pixel changes, compels the Defender to develop a robust and nuanced visual perception capable of withstanding complex visual interference.

Our primary contributions are threefold:
\begin{itemize}
    \item We propose a systematic data generation pipeline and will release the resulting structured dataset, \textbf{AOT-SFT}. This resource, composed of paired clean and adversarially manipulated high-resolution images, provides a unique tool for analyzing MLLM robustness and serves as the crucial bootstrapping corpus for our self-play framework.
    \item We propose \textbf{AOT: Adversarial Opponent Training}, a new self-play framework for MLLMs that autonomously generates training data through an iterative, adversarial process. This scalable approach directly targets and enhances the model's core perceptual abilities by co-evolving an image-editing Attacker and an MLLM Defender.
    \item Extensive experiments demonstrate that our \textbf{AOT} framework substantially enhances the perceptual robustness of MLLMs, outperforming strong baselines that rely on finite adversarial datasets, with notable benefits in reducing model hallucination and producing a highly transferable training curriculum.
\end{itemize}

\begin{figure*}[t!]
\centering
\includegraphics[width=\textwidth]{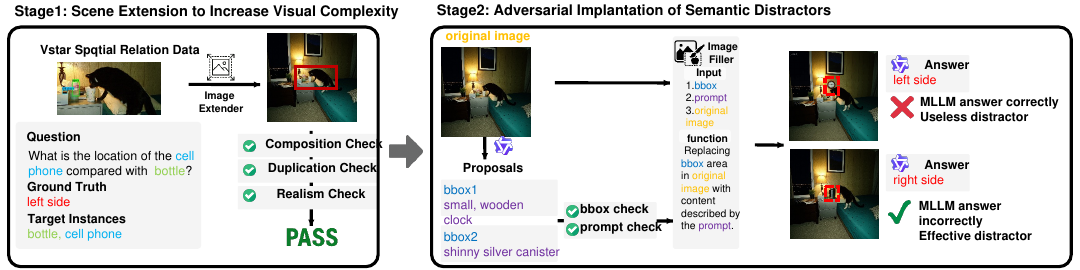}
\caption{An overview of our two-stage pipeline for generating the initial adversarial SFT dataset. \textbf{Stage 1: Scene Extension to Increase Visual Complexity.} We begin with a source image and apply outpainting to expand the scene, thereby increasing its complexity. The resulting image is then subjected to a rigorous filtering process, including Composition, Duplication, and Realism Checks, to ensure its quality and logical consistency with the original task. \textbf{Stage 2: Adversarial Implantation of Semantic Distractors.} For an extended image that a target MLLM can correctly interpret, we generate proposals for semantic distractors—new objects and their locations—to be inpainted into the scene. These proposals are validated to prevent spatial overlap or semantic duplication of existing target instances. \textbf{The final image is added to the SFT dataset only if the implanted distractor is effective, causing the MLLM to fail.}}
\label{fig:sft_pipeline}
\end{figure*}

\section{Related Work}
\label{sec:related work}

\subsection{Self-Play in Language Models}
Self-play has emerged as a key paradigm to overcome data bottlenecks in large language models \cite{wang2023self, Pourcel2025SelfImprovingLM, Zhou2025SelfChallengingLM}. These frameworks create dynamic learning environments where models improve through mutual interaction \cite{chen2023selfdebug, li2024autoif}, proving especially effective in domains with objective correctness criteria like code generation \cite{roziere2023code, luo2023wizardcoder, wei2024magicoder, le2022rltf, ni2023teaching}. To prevent the learning process from stagnating, advanced self-play systems implement a co-evolutionary dynamic, where the capabilities of all interacting agents are continuously and reciprocally enhanced \cite{Lin2025LearningTS, Wang2025CoEvolvingLC, Fang2025SeRLSR, Zhao2025AbsoluteZR}. Our work is inspired by this co-evolutionary principle, translating the adversarial interaction from text-based domains to the perceptual challenges of fine-grained vision.

\subsection{Self-Play and Adversarial Training in Vision-Language Models}
Self-play is an emerging strategy in Vision-Language Models (VLMs) to mitigate the prohibitive cost of creating high-quality multimodal datasets \cite{krishna2017visual, grauman2022ego4d}. Prior works are either fundamentally cooperative, using games for data generation and reasoning \cite{konyushkova2025vision, wang2025vision, chen2025g1}, or adversarial but focused on security vulnerabilities against multimodal jailbreaks \cite{dai2025secure}. Our work diverges from these by establishing a purely perceptual adversarial framework designed to remedy core robustness weaknesses. This distinguishes our approach from methods that rely on finite benchmarks with predefined distractions~\cite{liuRobustnessMultimodalLanguage2025} or scripted transformations for evaluation~\cite{yangDynamicMultimodalEvaluation2024b}. Critically, our framework introduces a co-evolution where the attacker autonomously discovers novel visual attacks by adapting to the defender, a dynamic learning process absent in prior work.

\section{Dataset}
\label{sec:data_generation}

\begin{figure}[t!]
\centering
\includegraphics[width=0.9\linewidth]{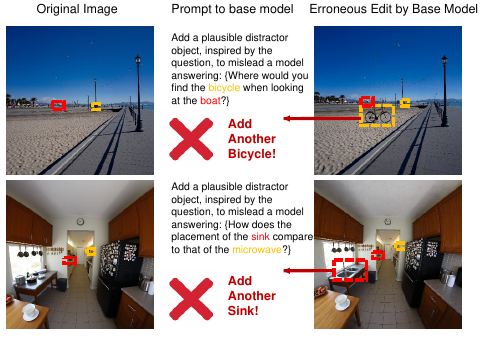}
\caption{Demonstration of the necessity for initial SFT. The base Qwen-Image-Edit model (without SFT), when prompted to generate a semantic distractor, fundamentally misunderstands the instruction. \textbf{Instead of adding a confusing element, it directly inserts the object mentioned in the question} (e.g., adding a bicycle when the question is about a bicycle's location relative to a boat). This behavior highlights its inability to comprehend complex, adversarial instructions, motivating our creation of an initial SFT dataset to explicitly teach this capability.}
\label{fig:sft_necessity}
\end{figure}

Our self-play framework necessitates an initial supervised fine-tuning (SFT) phase for the attacker model to address the cold-start problem. Off-the-shelf image editing models lack the nuanced capability to generate effective semantic distractors, often misunderstanding instructions and inserting the very objects mentioned in the question, thereby invalidating the task, illustrated in \Cref{fig:sft_necessity}. To overcome this, we designed a dedicated two-stage data generation pipeline, illustrated in \Cref{fig:sft_pipeline}, to create a high-quality dataset for this initial training. The resulting dataset, which we term \textbf{AOT-SFT}, is composed of triplets \((I', Q, I_{adv})\), where \(I'\) is a clean image, \(Q\) is the associated question, and \(I_{adv}\) is its adversarial counterpart.

\subsection{Data Generation Pipeline}
\label{sec:pipeline_details}

\textbf{Stage 1: Scene Extension to Increase Visual Complexity.}~~Our process begins with images from the VStar spatial relation dataset \cite{wu2023V*Bench}. To increase the visual complexity and provide more opportunities for adversarial manipulation, we first perform outpainting. We employ Qwen2.5-VL (72B) \cite{bai_qwen25-vl_2025} to generate a detailed descriptive prompt for each scene, which then guides an image extender model, OneReward \cite{gong2025onereward}, to expand the image canvas. A critical negative constraint is included in the prompt to forbid the generation of objects identical to the target instances mentioned in the original question, thereby preserving the ground-truth answer. The extended images then undergo a rigorous three-part filtering process:

(1) \textit{Composition Check}: To filter out composite images made from disparate scenes, we prompt Qwen2.5-VL (72B) to classify whether the image appears to be a consistent scene or a composition. Samples are discarded if the model's confidence in the ``composition'' class exceeds 50\%.

(2) \textit{Duplication Check}: This check ensures that the outpainting process has not inadvertently introduced new instances of the target objects, which would alter the validity of the original question-answer pair. We instruct Qwen2.5-VL (72B) to output the bounding boxes of all target instances. If the number of detected instances exceeds the original count, the image is filtered out.

(3) \textit{Realism Check}: To ensure the generated content is plausible, we again use Qwen2.5-VL (72B) to perform a binary classification of the scene's realism. Images flagged as unrealistic with a confidence above 50\% are discarded.

Images that successfully pass all three checks proceed to the next stage.

\textbf{Stage 2: Adversarial Implantation of Semantic Distractors.}~~The goal of this stage is to strategically insert distractor objects into the validated images from Stage 1. First, we form a clean candidate pool, \(D_{clean}\), by selecting only those extended images that the initial defender MLLM, \(M_{def}^{(0)}\), can answer correctly. This ensures our subsequent attacks target the model's robust knowledge rather than its pre-existing failures.

For each clean sample \((I', Q, A) \in D_{clean}\), we again leverage Qwen2.5-VL (72B) to generate a set of proposals for adversarial distractors. Each proposal consists of a bounding box (\(B\)) for inpainting and a textual description (\(P_{obj}\)) of the object to be implanted. These proposals are subject to two crucial integrity checks before execution:

(1) \textit{Bbox Overlap Check}: We verify that the proposed bounding box \(B\) does not spatially overlap with the bounding boxes of the existing target instances in \(I'\). This prevents the distractor from directly occluding or replacing the objects relevant to the question, which would change the problem itself.

(2) \textit{Description Check}: We parse the description prompt \(P_{obj}\) to ensure it does not contain keywords of the target instances. This prevents the model from generating another instance of the same object class, maintaining the logical integrity of the original question.

Proposals that pass both checks are then passed to an image filler model (OneReward), which performs inpainting to implant the object described by \(P_{obj}\) within the region defined by \(B\), producing a candidate adversarial image \(I_{adv}\). Finally, we determine the efficacy of the implanted distractor. The pair \((I_{adv}, Q)\) is fed to the initial defender model \(M_{def}^{(0)}\). If the model's prediction is incorrect, the distractor is deemed effective, and the triplet \((I', Q, I_{adv})\) is added to our final dataset. If the model answers correctly, the distractor is considered ineffective, and the candidate is discarded.

\subsection{The AOT-SFT Dataset}
\label{sec:vstar_advhr_dataset}

This two-stage pipeline yields \textbf{AOT-SFT}, a structured dataset specifically designed for studying and enhancing MLLM robustness. A key characteristic of this dataset is its paired structure, where each clean high-resolution image \(I'\), which a baseline MLLM can correctly interpret, is matched with an adversarial counterpart \(I_{adv}\) containing a validated, effective semantic distractor.

In this work, we use the collection of successful adversarial triplets from this pipeline to form the initial supervised fine-tuning dataset, \(D_{SFT}\). This dataset is essential for bootstrapping our co-evolutionary framework, teaching the attacker model the foundational skill of generating semantically plausible and effective distractors. While serving this critical role in our AOT framework, the full AOT-SFT dataset will be released to the public to facilitate broader research into the perceptual vulnerabilities and robustness of MLLMs.

\begin{figure*}[t!]
\centering
\includegraphics[width=\textwidth]{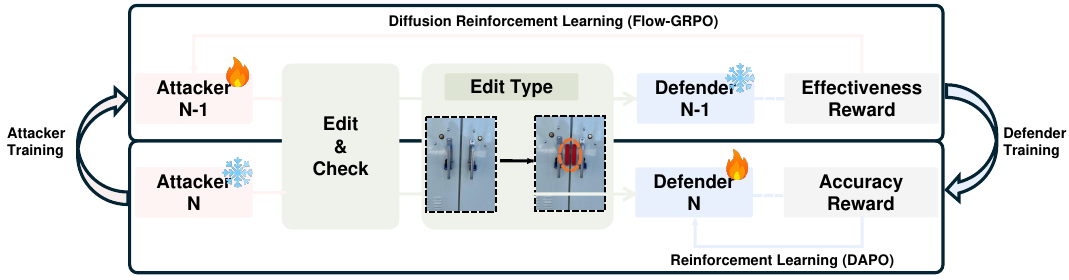}
\caption{An overview of our iterative attacker-defender co-evolution framework. The process consists of two interconnected training loops. \textbf{Attacker Evolution:} The active attacker (\(M_{atk}^{(N)}\)) is refined using Flow-GRPO. It learns to generate adversarial edits designed to deceive the frozen, previous-generation defender (\(M_{def}^{(N-1)}\)). The success of this deception provides the 'Effectiveness Reward' that drives the attacker's policy update. \textbf{Defender Enhancement:} Subsequently, the newly updated attacker generates challenging examples to train the active defender (\(M_{def}^{(N)}\)). The defender is updated via DAPO based on an 'Accuracy Reward' derived from its performance on these adversarial inputs. This cycle repeats, progressively enhancing the capabilities of both models.}
\label{fig:co_evolution}
\end{figure*}

\section{Iterative Attacker-Defender Co-evolution}
\label{sec:co_evolution}

Following the initial bootstrapping phase, we initiate an iterative co-evolutionary process designed to mutually enhance the capabilities of an attacker model and a defender model. This adversarial game, illustrated in \Cref{fig:co_evolution}, is orchestrated by policy optimization algorithms, creating a dynamic training environment where the defender's robustness is progressively strengthened against an ever-evolving threat landscape. The framework features an \textbf{Attacker} (\(M_{atk}\)), an image editing model tasked with generating semantic adversarial examples, and a \textbf{Defender} (\(M_{def}\)), the LVLM we aim to make more robust. The process unfolds over several rounds of alternating training, as detailed below.

\subsection{Attacker Evolution}
\label{sec:attacker_evolution}

In each round \(i\), the primary objective is to refine the attacker \(M_{atk}^{(i-1)}\) to generate more potent adversarial examples that can successfully deceive the current defender, \(M_{def}^{(i-1)}\). The attacker, specifically the Qwen Image Edit model, is trained using Flow-GRPO \cite{liu2025flow}, a policy optimization algorithm tailored for generative models. For this iterative process, we utilize a distinct subset of 300 samples, denoted \(D_{clean}^{(i)}\), drawn without replacement from our main clean dataset \(D_{clean}\). The crux of this training phase lies in the design of a composite reward function, \(R_{atk}\), that balances two critical objectives: maintaining the semantic integrity of the original scene and maximizing the adversarial impact.

\par\noindent\textbf{Semantic Integrity via Localized SSIM.}
A fundamental constraint for a valid adversarial attack is that the perturbation must not alter the core objects or relations relevant to the original question. To enforce this, we move beyond a global image comparison and implement a strict, localized integrity check. Let \(B_Q = \{b_1, b_2, \dots, b_k\}\) denote the set of bounding boxes corresponding to the critical objects mentioned in question \(Q\). For each generated image \(I_{adv}\), we compute the Structural Similarity Index Measure (SSIM) individually for each region defined by a bounding box \(b_j \in B_Q\), comparing the patch in \(I_{adv}\) with the corresponding patch in the original image \(I'\). An attack is considered to have corrupted the scene's premise if the SSIM score for any of these critical regions falls below a predefined threshold \(\tau_{ssim}\). Such an attempt is immediately assigned a reward of zero, preventing the model from learning to produce trivial or invalid edits. In our experiments, this stringent per-region check leads to the rejection of a notable fraction of invalid attack candidates, ensuring the quality of the adversarial examples.

\par\noindent\textbf{Adversarial Efficacy Reward.}\  For attacks that pass the SSIM check, the reward is determined by their effectiveness in fooling the defender. To penalize brittle attacks that might succeed by chance, we introduce a high-confidence failure criterion: a successful attack must cause the defender to produce an incorrect answer across two consecutive inference attempts under deterministic decoding (i.e., temperature set to 0). An attack that passes the integrity check receives a base reward of 0.2. This small reward incentivizes the model to explore valid edits even if they are not immediately successful. The full reward of 1.0 is only granted if the attack is both valid (passes SSIM) and consistently effective (causes two consecutive failures).
The composite reward function \(R_{atk}\) for a generated image \(I_{adv}\), given an original image \(I'\) and question \(Q\) with associated critical bounding boxes \(B_Q\), is formally defined as:
\begin{equation}
\label{eq:reward_atk}
R_{atk} = 
\begin{cases}
  1.0 & \text{if } \begin{subarray}{l} \min_{b_j \in B_Q} \text{SSIM}(I_{adv}[b_j], I'[b_j]) \ge \tau_{ssim} \text{ and} \\ M_{def}^{(i-1)}(I_{adv}, Q) \neq A \text{ (twice)} \end{subarray} \\
  0.2 & \text{if } \begin{subarray}{l} \min_{b_j \in B_Q} \text{SSIM}(I_{adv}[b_j], I'[b_j]) \ge \tau_{ssim} \text{ and} \\ M_{def}^{(i-1)}(I_{adv}, Q) = A \text{ (at least once)} \end{subarray} \\
  0   & \text{if } \min_{b_j \in B_Q} \text{SSIM}(I_{adv}[b_j], I'[b_j]) < \tau_{ssim}
\end{cases}
\end{equation}
where \(A\) is the ground-truth answer and \(I[b_j]\) denotes the image patch within bounding box \(b_j\). With this reward signal, we update the attacker's policy \(M_{atk}^{(i-1)}\) using the Flow-GRPO loss to obtain the next-generation attacker \(M_{atk}^{(i)}\).

\par\noindent\textbf{Curating the Adversarial Training Set.} Upon obtaining the updated attacker \(M_{atk}^{(i)}\), we deploy it on the current data subset \(D_{clean}^{(i)}\) to generate a large pool of adversarial candidates. To curate a high-quality training set for the next defender, we introduce a sophisticated filtering mechanism designed to identify examples that are challenging yet learnable. Each candidate \(I_{adv}\) is evaluated against the frozen, previous-generation defender \(M_{def}^{(i-1)}\) over ten separate inference trials using stochastic sampling (temperature set to 1.0). An adversarial example is selected for the defender's training set, \(D_{adv}^{(i)}\), only if the defender provides the correct answer between three and seven times, inclusive. This filtering strategy discards examples that are either too easy (accuracy $>$ 70\%) or excessively difficult (accuracy $<$ 30\%), ensuring that the defender is trained on a diet of maximally informative samples that effectively probe its decision boundary.

\subsection{Defender Enhancement}
\label{sec:defender_enhancement}
Upon the convergence of the attacker \(M_{atk}^{(i)}\) in round \(i\), we leverage its enhanced capabilities to generate a new, more challenging adversarial dataset \(D_{adv}^{(i)}\). This is accomplished by deploying \(M_{atk}^{(i)}\) on the entire clean dataset \(D_{clean}\). The resulting dataset, denoted as \(D_{adv}^{(i)}\), represents the latest and most difficult attack vectors:
\begin{equation}
\label{eq:dadv_def_gen}
D_{adv}^{(i)} = \left\{ \bigl(M_{atk}^{(i)}(I',Q),\, Q,\, A\bigr)\ \middle|\ (I',Q,A)\in D_{clean} \right\}.
\end{equation}

The defender \(M_{def}^{(i-1)}\) is then fine-tuned on this curated dataset to learn resilience against these novel attacks. For this stage, we utilize the DAPO\cite{yu2025dapo, shao2024deepseekmath} algorithm. The reward function \(R_{def}\) for the defender is designed to encourage both correctness and proper formatting.
\begin{equation}
\label{eq:reward_def}
R_{def} = 
\begin{cases}
  1.0 & \text{if } \text{prediction} = A \land \text{format is correct} \\
  0.8 & \text{if } \text{prediction} = A \land \text{format is incorrect} \\
  0   & \text{if } \text{prediction} \neq A
\end{cases}
\end{equation}
A primary reward of 0.8 is awarded if the model produces the correct answer \(A\), regardless of its output format. An additional bonus of 0.2 is given if the output also adheres to the expected format (e.g., providing the answer choice letter correctly), encouraging the model to maintain its instruction-following capabilities. The defender's policy \(M_{def}^{(i-1)}\) is then updated by minimizing the DAPO loss, yielding the more robust defender \(M_{def}^{(i)}\). This iterative cycle of attacker evolution and defender enhancement repeats, progressively improving the robustness of the defender against increasingly sophisticated semantic attacks.

\begin{table*}[t]
\centering
\footnotesize
\setlength{\tabcolsep}{1.5pt}
\caption{Main results on perceptual robustness and hallucination benchmarks. We report the Overall accuracy for \textbf{VStar} and \textbf{HRBench-4K/8K}, along with detailed sub-metrics. Our co-evolutionary framework is compared against the base model, an ablation, and several data augmentation baselines. The peak performance for each metric is highlighted in \textbf{bold}. Gains (\textcolor{red!80!black}{\(\Delta\)}) are calculated over the base model.}
\label{tab:main_results_combined}
\begin{tabular}{@{}l ccccc ccc c c c@{}}
\toprule
\multirow{2}{*}{\textbf{Method}} & \multicolumn{5}{c}{\textbf{VStar}} & \multicolumn{3}{c}{\textbf{HRBench-4K}} & \textbf{HRBench-8K} & \textbf{\makecell{Hallusion\\Bench}} & \textbf{POPE} \\
\cmidrule(lr){2-6} \cmidrule(lr){7-9} \cmidrule(lr){10-10} \cmidrule(lr){11-11} \cmidrule(lr){12-12}
& \textbf{Overall} & \textbf{G4V-H} & \textbf{OCR} & \textbf{D-Attr} & \textbf{R-Pos} & \textbf{Overall} & \textbf{S-obj} & \textbf{C-obj} & \textbf{Overall} & \textbf{aAcc} & \textbf{F1-Score} \\
\midrule
Base (\(M_{def}^{(0)}\)) & 71.01 & 82.35 & 100.0 & 60.00 & 73.68 & 64.12 & 79.50 & 48.75 & 64.88 & 67.51 & 77.12 \\
Ablation (Clean Data)  & 74.79 & 82.35 & 100.0 & 67.83 & 73.68 & 71.75 & 85.25 & 58.25 & 70.62 & 67.93 & 79.75 \\
\midrule
\multicolumn{12}{@{}l}{\textit{Training with finite Augmentation Datasets}} \\
Yang et al. \cite{yang2024dynamic} & 70.17 & 76.47 & 100.0 & 60.87 & 71.05 & 64.88 & 77.50 & 52.25 & 64.88 & 68.35 & 77.87 \\
Liu et al. \cite{liu2025robustness} (Add) & 73.11 & 94.12 & 100.0 & 64.35 & 71.05 & 66.88 & 79.50 & 54.25 & 66.25 & 67.61 & 78.10 \\
Liu et al. \cite{liu2025robustness} (Insert) & 74.79 & 82.35 & 100.0 & 66.96 & 75.00 & 67.12 & 81.75 & 52.50 & 66.13 & 68.45 & 79.09 \\
Liu et al. \cite{liu2025robustness} (All) & 76.05 & 82.35 & 100.0 & 70.43 & 73.68 & 67.12 & 83.25 & 51.00 & 66.75 & 67.61 & 78.15 \\
\midrule
\multicolumn{12}{@{}l}{\textit{Our Method (Iterative Co-evolution)}} \\
Defender Iter. 1 & 75.63 & 88.24 & 100.0 & 68.70 & 73.68 & 71.38 & 85.25 & 57.50 & 69.38 & 67.82 & 80.07 \\
Defender Iter. 2 & 77.31 & 94.12 & 100.0 & 70.43 & 75.00 & 72.00 & \textbf{84.75} & 59.25 & 69.88 & 68.24 & \textbf{80.35} \\
Defender Iter. 3 & \textbf{80.25} \textcolor{red!80!black}{(+9.24)} & \textbf{94.12} & \textbf{100.0} & \textbf{73.91} & \textbf{78.95} & \textbf{72.38} \textcolor{red!80!black}{(+8.26)} & 83.75 & \textbf{61.00} & \textbf{71.50} \textcolor{red!80!black}{(+6.62)} & \textbf{69.19} \textcolor{red!80!black}{(+1.68)} & 80.00 \textcolor{red!80!black}{(+2.88)} \\
\bottomrule
\end{tabular}
\end{table*}

\begin{table*}[t]
\centering
\small
\setlength{\tabcolsep}{5pt}
\caption{Performance on general multi-modal benchmarks. We evaluate the models on five diverse datasets covering multidisciplinary knowledge and reasoning (MMMU), vision-indispensable multimodal capability (MMStar), real-world understanding (RealWorldQA), multi-image visual perception (BLINK), and diagram understanding (AI2D). The results demonstrate that our robustness training framework maintains or improves general capabilities compared to the Base model. Gains (\textcolor{red!80!black}{\(\Delta\)}) are calculated over the base model.}
\label{tab:general_benchmarks}
\begin{tabular}{@{}l ccccc@{}}
\toprule
\textbf{Method} & \textbf{MMMU} (Dev) & \textbf{MMStar} & \textbf{RealWorldQA} & \textbf{BLINK} & \textbf{AI2D} \\
\midrule
Base (\(M_{def}^{(0)}\)) & 20.67 & 60.33 & 67.71 & 54.08 & 80.96 \\
Ablation (Clean Data) & 21.33 & 60.93 & 69.28 & 54.55 & 80.73 \\
\midrule
\multicolumn{6}{@{}l}{\textit{Training with finite Augmentation Datasets}} \\
Yang et al. \cite{yang2024dynamic} & 22.67 & 60.87 & 68.24 & 54.13 & 80.86 \\
Liu et al. \cite{liu2025robustness} (Add) & 19.33 & 60.40 & 67.97 & 54.28 & 80.70 \\
Liu et al. \cite{liu2025robustness} (Insert) & 20.67 & 61.20 & 68.24 & 54.02 & 80.86 \\
Liu et al. \cite{liu2025robustness} (All) & 18.00 & 60.80 & 68.76 & 53.50 & 80.63 \\
\midrule
\multicolumn{6}{@{}l}{\textit{Our Method (Iterative Co-evolution)}} \\
Defender Iter. 1 & 20.00 & 61.27 & 69.28 & 54.55 & 81.09 \\
Defender Iter. 2 & 23.33 & 61.40 & 69.28 & 54.76 & 81.25 \\
Defender Iter. 3 & \textbf{25.33} \textcolor{red!80!black}{(+4.66)} & \textbf{61.53} \textcolor{red!80!black}{(+1.20)} & \textbf{70.07} \textcolor{red!80!black}{(+2.36)} & \textbf{55.92} \textcolor{red!80!black}{(+1.84)} & \textbf{81.35} \textcolor{red!80!black}{(+0.39)} \\
\bottomrule
\end{tabular}
\end{table*}

\section{Experiments}
\label{sec:experiments}
\subsection{Experimental Setup}

\paragraph{Models.} Our framework features a co-evolutionary dynamic between a Defender model \(M_{def}\), whose robustness we enhance, and an Attacker model \(M_{atk}\) that generates visual perturbations. For the Defender (\(M_{def}\)), we employ Qwen2.5-VL (7B) as the primary base model. To assess generalization, we also apply the generated data to a diverse set of models, including two from the Qwen3-VL\cite{team2025qwen3} series and three from the Gemma-3\cite{team2025gemma} series. The Attacker (\(M_{atk}\)) is Qwen-Image-Edit, initialized via supervised fine-tuning and subsequently evolved through our proposed policy optimization framework.

\paragraph{Evaluation Benchmarks.}
We evaluate our method on three distinct categories of benchmarks to ensure a holistic assessment:
\begin{enumerate}
    \item \textbf{Perceptual Robustness:} We employ VStar \cite{wu2023V*Bench} for fine-grained spatial perception and HRBench \cite{wang2025divide} (both 4K and 8K variants) to test zero-shot generalization in high-resolution, information-dense scenarios.
    \item \textbf{Factuality and Hallucination:} To measure the reduction of visual illusions and language hallucinations, we use POPE \cite{li2023evaluating} (object presence) and HallusionBench \cite{guan2024hallusionbench}.
    \item \textbf{General Capabilities:} To verify that our robustness training does not degrade general performance, we evaluate on MMMU \cite{yue2024mmmu}, MMStar \cite{chen2024we}, RealWorldQA\cite{xai2024realworldqa}, BLINK\cite{fu2024blink}, and AI2D \cite{kembhavi2016diagram}.
\end{enumerate}

\paragraph{Implementation Details.} Our co-evolutionary framework is conducted for three iterations. For the attacker training, we fine-tune the Qwen-Image-Edit model using a fast variant of Flow-GRPO with LoRA. The diffusion process uses 40 steps for both training and inference, with key hyperparameters for the fast variant including a noise level of 1.5 and a Stochastic Differential Equation (SDE) window size of 4. The SSIM threshold for the integrity check, \(\tau_{ssim}\), is set to 0.3 based on empirical validation. For the defender training, we fine-tune the Qwen2.5-VL(7B) model using the DAPO algorithm with a learning rate of \(1 \times 10^{-6}\) and a global batch size of 512. The defender's training in each iteration is stopped when its policy converges, indicated by the model consistently achieving maximum (1.0) or minimum (0.0) rewards across the training batch.

\paragraph{Baselines.} 
To provide a comprehensive evaluation, we compare our co-evolutionary framework against several baselines derived from recent data augmentation works. We include a method based on the ``hard'' finite datasets (LLaVABench-Hard, MMVet-Hard) from \cite{yang2024dynamic}. Additionally, we evaluate three variants using the distraction-based data from \cite{liu2025robustness}: training on ``add'' visual distractions, ``insert'' visual distractions, and a combined ``All'' dataset. For a fair comparison, all baseline models are fine-tuned from the same base model using the identical DAPO optimization protocol as our defender.

\subsection{Main Results on Perceptual Robustness}

The comprehensive results for perceptual robustness are presented in \cref{tab:main_results_combined}. Our proposed AOT framework effectively enhances the perceptual robustness of MLLMs across multiple resolutions. After three iterations, our Defender model achieves 80.25\% on VStar and 72.38\% on HRBench-4K, representing significant gains of \textbf{+9.24} and \textbf{+8.26} points over the base model, respectively. Notably, on the more challenging HRBench-8K, our method improves performance from 64.88\% to 71.50\%, demonstrating that the robustness gains generalize well to ultra-high-resolution inputs.

The model's performance improves progressively with each co-evolutionary iteration. On VStar, the Defender's score increases from 71.01\% to 75.63\% (Iter. 1), 77.31\% (Iter. 2), and finally 80.25\% (Iter. 3). This consistent growth highlights the benefit of the co-evolutionary dynamic, where the Defender learns from an increasingly challenging curriculum generated by the evolving Attacker. Our dynamic approach also outperforms baselines trained on finite adversarial datasets, surpassing the strongest distraction-based baseline (Liu et al. (All)) by +4.20 points on VStar and +4.75 points on HRBench-8K.

\subsection{Impact on Model Hallucination}

We test the hypothesis that enhanced perceptual robustness concurrently reduces hallucination. As detailed in \cref{tab:main_results_combined}, our method improves hallucination metrics, with our final Defender increasing the POPE F1-score by \textbf{+2.88} points and the HallusionBench aAcc by \textbf{+1.68} points over the base model. Our model also outperforms the data augmentation baselines on hallucination metrics. For instance, our model's POPE F1-score of 80.00 is higher than the 78.15 achieved by the strongest baseline (Liu et al. (All)). This suggests our framework enhances core perceptual acuity, which better grounds the model's responses in visual evidence and improves factuality.

\subsection{Impact on General Capabilities}

A common concern with adversarial or robustness-oriented fine-tuning is the potential for catastrophic forgetting, where the model loses its general capabilities. We address this by evaluating our model on a suite of five general multi-modal benchmarks, as shown in \cref{tab:general_benchmarks}.

The results indicate that our framework not only preserves but often improves general performance. On MMMU, a challenging benchmark for expert reasoning, our final model achieves 25.33\%, surpassing both the Base model (20.67\%) and the Ablation with clean data (21.33\%). Similarly, on benchmarks requiring fine-grained visual attention such as BLINK and RealWorldQA, our method achieves steady gains (e.g., +2.36 points on RealWorldQA). 

In contrast, the baselines exhibit mixed results. While the baseline trained on clean data shows slight improvements, the distraction-based methods (Liu et al.) often suffer from performance degradation on complex reasoning tasks like MMMU (dropping to 18.00\% in the ``All'' setting). This suggests that randomly adding distractions without a guided curriculum may disrupt the model's internal representations. Our co-evolutionary approach, by providing a tailored and progressive difficulty curve, allows the model to internalize robustness as a generalized visual skill rather than overfitting to specific noise patterns, thereby supporting broader multi-modal competencies.

\subsection{Ablation Study on Curriculum Selection}
\label{sec:curriculum_ablation}
To demonstrate the effectiveness of our curriculum selection strategy, we conduct an ablation study. We show that a curriculum with balanced difficulty and semantic integrity is crucial for optimal performance. As shown in \cref{tab:curriculum_ablation}, our proposed method significantly outperforms variants trained on poorly structured curricula.

\begin{table}[t]
\centering
\small
\setlength{\tabcolsep}{5.5pt}
\caption{Ablation study on our curriculum selection strategy. Our full method, using a balanced curriculum with an integrity check, achieves the best performance. Performance degradation (\textcolor{red!80!black}{\(\Delta\)}) compared to our method is shown in red.}
\label{tab:curriculum_ablation}
\begin{tabular}{@{}lcc@{}}
\toprule
\textbf{Curriculum Details} & \textbf{VStar} & \textbf{HRBench-4K} \\
\midrule
\textbf{\makecell[l]{Our Method\\(Balanced w/ Integrity Check)}} & \textbf{77.31} & \textbf{72.00} \\
\midrule
\multicolumn{3}{@{}l}{\textit{Ablations on Difficulty Range}} \\
\quad All Adversarial Examples & 76.47 \textcolor{red!80!black}{(-0.84)} & 70.75 \textcolor{red!80!black}{(-1.25)} \\
\quad Hardest Examples Only & 73.11 \textcolor{red!80!black}{(-4.20)} & 69.88 \textcolor{red!80!black}{(-2.12)} \\
\quad Easiest Examples Only & 76.47 \textcolor{red!80!black}{(-0.84)} & 71.00 \textcolor{red!80!black}{(-1.00)} \\
\midrule
\multicolumn{3}{@{}l}{\textit{Ablation on Data Integrity}} \\
\quad w/o Integrity Check (SSIM) & 77.31 \textcolor{red!80!black}{(-0.00)} & 70.75 \textcolor{red!80!black}{(-1.25)} \\
\bottomrule
\end{tabular}
\end{table}

The results in \cref{tab:curriculum_ablation} clearly validate our design. Focusing on the difficulty range, we observe that training on only the hardest samples significantly harms performance, dropping VStar accuracy by more than 4 points. This confirms that an excessively challenging curriculum can be detrimental. While training on all available adversarial examples yields reasonable performance, our balanced approach provides a clear advantage by focusing the model on the most informative samples.

The semantic integrity of the curriculum is also critical. Removing the SSIM check results in a noticeable performance drop on HRBench-4K, demonstrating its necessity. While the VStar score remains unaffected, we attribute this to the attacker's SSIM-based reward which already disincentivizes severe semantic corruption. Nonetheless, the explicit filter is vital for ensuring consistent robustness.

\begin{table}[t]
\centering
\caption{Generalization of the iterative adversarial curriculum. The curriculum, generated with Qwen2.5-VL 7B, is applied to various models. The gain (\textcolor{red!80!black}{\(\Delta\)}) is calculated over the \textbf{Base Model}. The data shows consistent, transferable improvements across model families and scales.}
\label{tab:generalization_iterative_compact}
\small
\begin{tabular}{@{}l cc@{}}
\toprule
\textbf{Method} & \textbf{VStar} & \textbf{HRBench-4K} \\
\midrule
\multicolumn{3}{l}{\textbf{Qwen3-VL (4B)}} \\
\quad Base Model & 73.11 & 70.62 \\
\quad + Clean Data (Baseline) & 77.31 & 74.75 \\
\quad + Curriculum - Iter 1 & 80.25 & 76.25 \\
\quad + Curriculum - Iter 2 & 79.41 & 76.38 \\
\quad + Curriculum - Iter 3 & 81.51 \textbf{\textcolor{red!80!black}{(+8.40)}} & 77.12 \textbf{\textcolor{red!80!black}{(+6.50)}} \\
\midrule
\multicolumn{3}{l}{\textbf{Qwen3-VL (8B)}} \\
\quad Base Model & 78.99 & 70.25 \\
\quad + Clean Data (Baseline) & 81.93 & 75.88 \\
\quad + Curriculum - Iter 1 & 83.61 & 77.25 \\
\quad + Curriculum - Iter 2 & 84.03 & 76.50 \\
\quad + Curriculum - Iter 3 & 86.55 \textbf{\textcolor{red!80!black}{(+7.56)}} & 77.50 \textbf{\textcolor{red!80!black}{(+7.25)}} \\
\midrule
\multicolumn{3}{l}{\textbf{Gemma-3-IT (4B)}} \\
\quad Base Model & 37.82 & 46.62 \\
\quad + Clean Data (Baseline) & 38.24 & 46.75 \\
\quad + Curriculum - Iter 1 & 41.60 & 47.50 \\
\quad + Curriculum - Iter 2 & 42.02 & 48.75 \\
\quad + Curriculum - Iter 3 & 42.44 \textbf{\textcolor{red!80!black}{(+4.62)}} & 50.75 \textbf{\textcolor{red!80!black}{(+4.13)}} \\
\midrule
\multicolumn{3}{l}{\textbf{Gemma-3-IT (27B)}} \\
\quad Base Model & 51.26 & 57.88 \\
\quad + Clean Data (Baseline) & 53.78 & 60.12 \\
\quad + Curriculum - Iter 1 & 53.36 & 61.00 \\
\quad + Curriculum - Iter 2 & 56.72 & 62.00 \\
\quad + Curriculum - Iter 3 & 55.88 \textbf{\textcolor{red!80!black}{(+4.62)}} & 63.25 \textbf{\textcolor{red!80!black}{(+5.37)}} \\
\bottomrule
\end{tabular}
\end{table}

\subsection{Generalizability of the Adversarial Curriculum}

\begin{figure*}[t!]
\centering
\includegraphics[width=\textwidth]{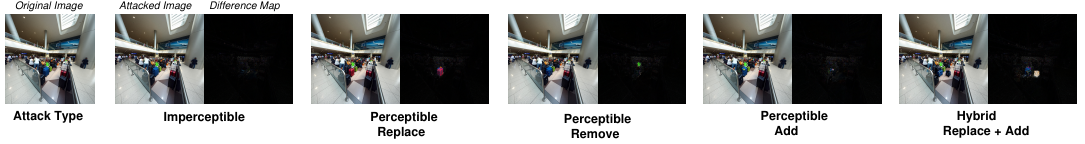}
\caption{Qualitative examples of the diverse attack strategies autonomously discovered by our attacker model. \textbf{Crucially, these strategies generalized far beyond our initial SFT dataset, which only contained examples of object addition.} The figure displays the original image (far left) followed by the results of five distinct attack types. These include \textbf{imperceptible pixel-level perturbations} and four types of perceptible semantic manipulations: \textbf{object replacement} (e.g., changing a blue suitcase to red), \textbf{object removal} (e.g., a green bag), \textbf{object addition} (e.g., a green tag), and a \textbf{hybrid attack} that combines multiple manipulations. For each attack, we show the attacked image and a difference map highlighting the manipulated regions.}
\label{fig:attack_analysis}
\end{figure*}

To evaluate the transferability of our method, we applied the iteratively generated adversarial curriculum to a wide range of MLLMs. As shown in \cref{tab:generalization_iterative_compact}, the curriculum yields substantial improvements across different model architectures and scales, demonstrating its broad applicability. Crucially, the performance gains generally surpass not only the base models but also a strong baseline fine-tuned on clean data, underscoring the unique value of the adversarially generated examples.

The curriculum demonstrates strong intra-family and cross-family generalization. For instance, the final iteration boosts Qwen3-VL (8B) by \textbf{+7.56} points on VStar over the base model, significantly outperforming the clean baseline. This capability transfers effectively to other architectures, enhancing Gemma-3-IT (27B) by a significant \textbf{+5.37} points on HRBench-4K. While observing some variance, the overall results show a clear trend of progressive improvement with each training iteration. This pattern of steady gains confirms that our co-evolutionary framework produces a broadly effective curriculum that enhances fundamental perceptual skills.

\subsection{Analysis of Emergent Attacker Strategies}

A key outcome of our co-evolutionary framework is the attacker's ability to autonomously discover a diverse repertoire of attack strategies, despite being initialized with only object addition examples. As illustrated in \cref{fig:attack_analysis}, the attacker learned to generate both imperceptible pixel-level perturbations and a variety of perceptible semantic manipulations. These emergent tactics include sophisticated strategies like \textbf{object replacement}, \textbf{object removal}, and \textbf{hybrid attacks}, which combine multiple manipulations. This autonomous discovery of a rich attack taxonomy is crucial; it creates a dynamic and varied training curriculum that prevents the defender from overfitting to a single threat type, thereby forcing it to develop a more holistic and robust perceptual system.

\section{Conclusion}
\label{sec:conclusion}
We presented a co-evolutionary framework that enhances the perceptual robustness of Multimodal Large Language Models through adversarial self-play. Our approach features an image-editing Attacker and a Defender MLLM that improve in tandem, with the Attacker generating a dynamic and challenging visual curriculum. Our experiments show this process yields a more robust Defender that achieves state-of-the-art performance on perceptual benchmarks and exhibits reduced hallucination, underscoring the benefits of our adaptive training paradigm. This underscores the superiority of an adaptive curriculum over finite adversarial datasets.

Future work could enhance the Attacker's generative diversity and explore more efficient co-evolutionary training schemes. However, our framework is currently centered on VQA tasks with objective correctness. Extending this self-play paradigm to open-ended generative tasks where deception criteria are subjective remains a significant challenge. Nevertheless, our work presents a scalable path away from finite datasets, marking a step towards building more resilient and reliable MLLMs.

{
    \small
    \bibliographystyle{ieeenat_fullname}
    \bibliography{main}

@String(IJCV  = {International Journal of Computer Vision})

@String(CVPR  = {Proceedings of the IEEE/CVF Conference on Computer Vision and Pattern Recognition (CVPR)})

@String(ECCV  = {Proceedings of the European Conference on Computer Vision (ECCV)})

@String(NIPS  = {Advances in Neural Information Processing Systems (NeurIPS)})

@String(AAAI  = {Proceedings of the AAAI Conference on Artificial Intelligence})

@String(ICML  = {International Conference on Machine Learning (ICML)})

@String(ACL   = {Proceedings of the Association for Computational Linguistics (ACL)})

@article{hurst2024gpt,
  title={{GPT}-4o System Card},
  author={Hurst, Aaron and Lerer, Adam and Goucher, Adam P and Perelman, Adam and Ramesh, Aditya and Clark, Aidan and Ostrow, AJ and Welihinda, Akila and Hayes, Alan and Radford, Alec and others},
  journal={arXiv preprint arXiv:2410.21276},
  year={2024}
}

@article{comanici2025gemini,
  title={{Gemini} 2.5: Pushing the Frontier with Advanced Reasoning, Multimodality, Long Context, and Next Generation Agentic Capabilities},
  author={Comanici, Gheorghe and Bieber, Eric and Schaekermann, Mike and Pasupat, Ice and Sachdeva, Noveen and Dhillon, Inderjit and Blistein, Marcel and Ram, Ori and Zhang, Dan and Rosen, Evan and others},
  journal={arXiv preprint arXiv:2507.06261},
  year={2025}
}

@article{guo2025deepseek,
  title={{DeepSeek-R1}: Incentivizing Reasoning Capability in {LLMs} via Reinforcement Learning},
  author={Guo, Daya and Yang, Dejian and Zhang, Haowei and Song, Junxiao and Zhang, Ruoyu and Xu, Runxin and Zhu, Qihao and Ma, Shirong and Wang, Peiyi and Bi, Xiao and others},
  journal={arXiv preprint arXiv:2501.12948},
  year={2025}
}

@article{yang2025qwen3,
  title={{Qwen3} Technical Report},
  author={Yang, An and Li, Anfeng and Yang, Baosong and Zhang, Beichen and Hui, Binyuan and Zheng, Bo and Yu, Bowen and Gao, Chang and Huang, Chengen and Lv, Chenxu and others},
  journal={arXiv preprint arXiv:2505.09388},
  year={2025}
}

@misc{llama3modelcard,
  title={{Llama 3} Model Card},
  author={{AI@Meta}},
  year={2024},
  howpublished={\url{https://github.com/meta-llama/llama3/blob/main/MODEL_CARD.md}}
}

@article{bai_qwen25-vl_2025,
  title={{Qwen2.5-VL} Technical Report},
  author={Bai, Shuai and Chen, Keqin and Liu, Xuejing and Wang, Jialin and Ge, Wenbin and Song, Sibo and Dang, Kai and Wang, Peng and Wang, Shijie and Tang, Jun and others},
  journal={arXiv preprint arXiv:2502.13923},
  year={2025}
}

@article{zhu_internvl3_2025,
  title={{InternVL3}: Exploring Advanced Training and Test-Time Recipes for Open-Source Multimodal Models},
  author={Zhu, Jinguo and Wang, Weiyun and Chen, Zhe and Liu, Zhaoyang and Ye, Shenglong and Gu, Lixin and Tian, Hao and Duan, Yuchen and Su, Weijie and Shao, Jie and others},
  journal={arXiv preprint arXiv:2504.10479},
  year={2025}
}

@inproceedings{liu2023llava,
  title={Visual Instruction Tuning},
  author={Liu, Haotian and Li, Chunyuan and Wu, Qingyang and Lee, Yong Jae},
  booktitle=NIPS,
  year={2023}
}

@article{usama2025analysing,
  title={Analysing the Robustness of {Vision-Language Models} to Common Corruptions},
  author={Usama, Muhammad and Asim, Syeda Aishah and Ali, Syed Bilal and Wasim, Syed Talal and Mansoor, Umair Bin},
  journal={arXiv preprint arXiv:2504.13690},
  year={2025}
}

@article{liu2025robustness,
  title={On the Robustness of Multimodal Language Model Towards Distractions},
  author={Liu, Ming and Chen, Hao and Wang, Jindong and Zhang, Wensheng},
  journal={arXiv preprint arXiv:2502.09818},
  year={2025}
}

@article{yang2024dynamic,
  title={Dynamic Multimodal Evaluation with Flexible Complexity by {Vision-Language} Bootstrapping},
  author={Yang, Yue and Zhang, Shuibai and Shao, Wenqi and Zhang, Kaipeng and Bin, Yi and Wang, Yu and Luo, Ping},
  journal={arXiv preprint arXiv:2410.08695},
  year={2024}
}

@inproceedings{chen2024spatialvlm,
  title={{SpatialVLM}: Endowing {Vision-Language Models} with Spatial Reasoning Capabilities},
  author={Chen, Boyuan and Xu, Zhuo and Kirmani, Sean and Ichter, Brain and Sadigh, Dorsa and Guibas, Leonidas and Xia, Fei},
  booktitle=CVPR,
  year={2024}
}

@article{chen2025spatial,
  title={Why is Spatial Reasoning Hard for {VLMs}? An Attention Mechanism Perspective on Focus Areas},
  author={Chen, Shiqi and Zhu, Tongyao and Zhou, Ruochen and Zhang, Jinghan and Gao, Siyang and Niebles, Juan Carlos and Geva, Mor and He, Junxian and Wu, Jiajun and Li, Manling},
  journal={arXiv preprint arXiv:2503.01773},
  year={2025}
}

@inproceedings{guo2024llava,
  title={{LLaVA-UHD}: An {LMM} Perceiving Any Aspect Ratio and High-Resolution Images},
  author={Guo, Zonghao and Xu, Ruyi and Yao, Yuan and Cui, Junbo and Ni, Zanlin and Ge, Chunjiang and Chua, Tat-Seng and Liu, Zhiyuan and Huang, Gao},
  booktitle=ECCV,
  year={2024}
}

@article{zhang2024beyond,
  title={Beyond {LLaVA-HD}: Diving into High-Resolution Large Multimodal Models},
  author={Zhang, Yi-Fan and Wen, Qingsong and Fu, Chaoyou and Wang, Xue and Zhang, Zhang and Wang, Liang and Jin, Rong},
  journal={arXiv preprint arXiv:2406.08487},
  year={2024}
}

@article{pattnayak2024survey,
  title={Survey of Large Multimodal Model Datasets, Application Categories and Taxonomy},
  author={Pattnayak, Priyaranjan and Patel, Hitesh Laxmichand and Kumar, Bhargava and Agarwal, Amit and Banerjee, Ishan and Panda, Srikant and Kumar, Tejaswini},
  journal={arXiv preprint arXiv:2412.17759},
  year={2024}
}

@article{liang2024foundations,
  title={Foundations \& Trends in Multimodal Machine Learning: Principles, Challenges, and Open Questions},
  author={Liang, Paul Pu and Zadeh, Amir and Morency, Louis-Philippe},
  journal={ACM Computing Surveys},
  volume={56},
  number={10},
  pages={1--42},
  year={2024}
}

@inproceedings{wu2023multimodal,
  title={Multimodal Large Language Models: A Survey},
  author={Wu, Jiayang and Gan, Wensheng and Chen, Zefeng and Wan, Shicheng and Yu, Philip S},
  booktitle={IEEE International Conference on Big Data (BigData)},
  pages={2247--2256},
  year={2023}
}

@article{villalobos2022will,
  title={Will We Run Out of Data? An Analysis of the Limits of Scaling Datasets in Machine Learning},
  author={Villalobos, Pablo and Sevilla, Jaime and Heim, Lennart and Besiroglu, Tamay and Hobbhahn, Marius and Ho, Anson},
  journal={arXiv preprint arXiv:2211.04325},
  year={2022}
}

@article{pan2025beyond,
  title={Beyond Benchmarks: Dynamic, Automatic and Systematic Red-Teaming Agents for Trustworthy Medical Language Models},
  author={Pan, Jiazhen and Jian, Bailiang and Hager, Paul and Zhang, Yundi and Liu, Che and Jungmann, Friedrike and Li, Hongwei Bran and You, Chenyu and Wu, Junde and Zhu, Jiayuan and others},
  journal={arXiv preprint arXiv:2508.00923},
  year={2025}
}

@inproceedings{rice2020overfitting,
  title={Overfitting in Adversarially Robust Deep Learning},
  author={Rice, Leslie and Wong, Eric and Kolter, Zico},
  booktitle=ICML,
  year={2020}
}

@inproceedings{stutz2020confidence,
  title={Confidence-Calibrated Adversarial Training: Generalizing to Unseen Attacks},
  author={Stutz, David and Hein, Matthias and Schiele, Bernt},
  booktitle=ICML,
  year={2020}
}

@article{yangDynamicMultimodalEvaluation2024b,
  title={Dynamic Multimodal Evaluation with Flexible Complexity by {Vision-Language} Bootstrapping},
  author={Yang, Yue and Zhang, Shuibai and Shao, Wenqi and Zhang, Kaipeng and Bin, Yi and Wang, Yu and Luo, Ping},
  journal={arXiv preprint arXiv:2410.08695},
  year={2024}
}

@article{liuRobustnessMultimodalLanguage2025,
  title={On the Robustness of Multimodal Language Model Towards Distractions},
  author={Liu, Ming and Chen, Hao and Wang, Jindong and Zhang, Wensheng},
  journal={arXiv preprint arXiv:2502.09818},
  year={2025}
}

@article{wu2023V*Bench,
  title={{V*}: Guided Visual Search as a Core Mechanism in Multimodal {LLMs}}, 
  author={Wu, Penghao and Xie, Saining},
  journal={arXiv preprint arXiv:2312.14135},
  year={2023}
}

@article{liu2025flow,
  title={{Flow-GRPO}: Training Flow Matching Models via Online {RL}},
  author={Liu, Jie and Liu, Gongye and Liang, Jiajun and Li, Yangguang and Liu, Jiaheng and Wang, Xintao and Wan, Pengfei and Zhang, Di and Ouyang, Wanli},
  journal={arXiv preprint arXiv:2505.05470},
  year={2025}
}

@article{chen2023selfdebug,
  title={{Self-Debug}: Autonomous Self-Debugging for Large Language Models in Code Generation},
  author={Chen, Tianlan and Zhang, Zhiye and Wang, Zelin and Liu, Yihong and Zhang, Shankai and Wang, Yuxiang and Lin, Bill Yuchen and Wang, Peidong and Yin, Bei and Lu, Yefan and others},
  journal={arXiv preprint arXiv:2304.05128},
  year={2023}
}

@article{Fang2025SeRLSR,
  title={{SeRL}: Self-Play Reinforcement Learning for Large Language Models with Limited Data},
  author={Wenkai Fang and Shunyu Liu and Yang Zhou and Kongcheng Zhang and Tongya Zheng and others},
  journal={arXiv preprint arXiv:2505.20347},
  year={2025}
}

@article{le2022rltf,
  title={{RLTF}: Reinforcement Learning from Unit Test Feedback},
  author={Le, Hieu Tran and Yuan, Yue and Allamanis, Miltiadis and Splash, Peter C. and Polozov, Oleksandr},
  journal={arXiv preprint arXiv:2207.14577},
  year={2022}
}

@article{li2024autoif,
  title={{AutoIF}: Aligning {LLMs} with Divergent Human Preferences through Iterative Self-Revised Instruction Following},
  author={Li, Haoran and Yuan, ZSEQH. and Guo, Ruomeng and Zhang, Yirui and Liu, Zhaoye and Sun, Maosong and Zhou, Jie},
  journal={arXiv preprint arXiv:2406.13542},
  year={2024}
}

@article{Lin2025LearningTS,
  title={Learning to Solve and Verify: A Self-Play Framework for Code and Test Generation},
  author={Zi Lin and Sheng Shen and Jingbo Shang and Jason Weston and Yixin Nie},
  journal={arXiv preprint arXiv:2502.14948},
  year={2025}
}

@article{luo2023wizardcoder,
  title={{WizardCoder}: Empowering Code Large Language Models with {Evol-Instruct}},
  author={Luo, Ziyang and Xu, Can and Zhao, Pu and Sun, Qingfeng and Geng, Xiubo and Hu, Wenxiang and Tao, Chongyang and Ma, Jing and Lin, Qingwei and Jiang, Daxin},
  journal={arXiv preprint arXiv:2306.08568},
  year={2023}
}

@inproceedings{ni2023teaching,
  title={Teaching Language Models to Rerank Code with Execution Feedback},
  author={Ni, Allen and Allal, Loubna Ben and Hao, Shuyang and Jain, Paras and Mu, Nan and Christopoulou, Martha and Peng, Cheng-Yu and Sutton, Charles and Nijkamp, Erik and Polozov, Oleksandr and others},
  booktitle=ICML,
  year={2023}
}

@article{Pourcel2025SelfImprovingLM,
  title={Self-improving language models for evolutionary program synthesis: A case study on ARC-AGI},
  author={Pourcel, Julien and Colas, C{\'e}dric and Oudeyer, Pierre-Yves},
  journal={arXiv preprint arXiv:2507.14172},
  year={2025}
}

@article{roziere2023code,
  title={{Code Llama}: Open Foundation Models for Code},
  author={Roziere, Baptiste and Gehring, Jonas and Gloeckle, Fabian and Sootla, Sten and Gat, Itai and Tan, Xiaoqing Ellen and Adi, Yossi and Liu, Jingyu and Sauvestre, Romain and Remez, Tal and others},
  journal={arXiv preprint arXiv:2308.12950},
  year={2023}
}

@article{Wang2025CoEvolvingLC,
  title={Co-Evolving {LLM} Coder and Unit Tester via Reinforcement Learning},
  author={Yinjie Wang and Ling Yang and Ye Tian and Ke Shen and Mengdi Wang},
  journal={arXiv preprint arXiv:2506.03136},
  year={2025}
}

@inproceedings{wang2023self,
  title={{Self-Instruct}: Aligning Language Models with Self-Generated Instructions},
  author={Wang, Yizhong and Kordi, Yeganeh and Mishra, Swaroop and Liu, Alisa and Smith, Noah A and Khashabi, Daniel and Hajishirzi, Hannaneh},
  booktitle=ACL,
  year={2023}
}

@inproceedings{wei2024magicoder,
  title={{Magicoder}: Empowering Code Generation with {OSS-Instruct}},
  author={Wei, Yuxiang and Wang, Zhe and Liu, Jiawei and Ding, Yifeng and Zhang, Lingming},
  booktitle=ICML,
  year={2024}
}

@article{Zhao2025AbsoluteZR,
  title={Absolute Zero: Reinforced Self-Play Reasoning with Zero Data},
  author={Andrew Zhao and Yiran Wu and Yang Yue and Tong Wu and Quentin Xu and others},
  journal={arXiv preprint arXiv:2505.03335},
  year={2025}
}

@article{Zhou2025SelfChallengingLM,
  title={Self-Challenging Language Model Agents},
  author={Yifei Zhou and Sergey Levine and Jason E. Weston and Xian Li and Sainbayar Sukhbaatar},
  journal={arXiv preprint arXiv:2506.01716},
  year={2025}
}

@article{krishna2017visual,
  title={{Visual Genome}: Connecting Language and Vision Using Crowdsourced Dense Image Annotations},
  author={Krishna, Ranjay and Zhu, Yuke and Groth, Oliver and Johnson, Justin and Hata, Kenji and Kravitz, Joshua and Chen, Stephanie and Kalantidis, Yannis and Li, Li-Jia and Shamma, David A and others},
  journal=IJCV,
  volume={123},
  number={1},
  pages={32--73},
  year={2017}
}

@inproceedings{grauman2022ego4d,
  title={{Ego4d}: Around the World in 3,000 Hours of Egocentric Video},
  author={Grauman, Kristen and Westbury, Andrew and Byrne, Eugene and Chavis, Zachary and Furnari, Antonino and Girdhar, Rohit and Hamburger, Jackson and Jiang, Hao and Liu, Miao and Liu, Xingyu and others},
  booktitle=CVPR,
  year={2022}
}

@article{konyushkova2025vision,
  title={{Vision-Language Model} Dialog Games for Self-Improvement},
  author={Konyushkova, Ksenia and Kaplanis, Christos and Cabi, Serkan and Denil, Misha},
  journal={arXiv preprint arXiv:2502.02740},
  year={2025}
}

@article{chen2025g1,
  title={{G1}: Bootstrapping Perception and Reasoning Abilities of {Vision-Language Model} via Reinforcement Learning},
  author={Chen, Liang and Gao, Hongcheng and Liu, Tianyu and Huang, Zhiqi and Sung, Flood and Zhou, Xinyu and Wu, Yuxin and Chang, Baobao},
  journal={arXiv preprint arXiv:2505.13426},
  year={2025}
}

@article{wang2025vision,
  title={{Vision-Zero}: Scalable {VLM} Self-Improvement via Strategic Gamified Self-Play},
  author={Wang, Qinsi and Liu, Bo and Zhou, Tianyi and Shi, Jing and Lin, Yueqian and Chen, Yiran and Li, Hai Helen and Wan, Kun and Zhao, Wentian},
  journal={arXiv preprint arXiv:2509.25541},
  year={2025}
}

@article{dai2025secure,
  title={{Secure Tug-of-War (SecTOW)}: Iterative Defense-Attack Training with Reinforcement Learning for Multimodal Model Security},
  author={Dai, Muzhi and Liu, Shixuan and Zhao, Zhiyuan and Gao, Junyu and Sun, Hao and Li, Xuelong},
  journal={arXiv preprint arXiv:2507.22037},
  year={2025}
}

@article{gong2025onereward,
  title={{OneReward}: Unified Mask-Guided Image Generation via Multi-Task Human Preference Learning},
  author={Gong, Yuan and Wang, Xionghui and Wu, Jie and Wang, Shiyin and Wang, Yitong and Wu, Xinglong},
  journal={arXiv preprint arXiv:2508.21066},
  year={2025}
}

@article{yu2025dapo,
  title={{DAPO}: An Open-Source {LLM} Reinforcement Learning System at Scale},
  author={Yu, Qiying and Zhang, Zheng and Zhu, Ruofei and Yuan, Yufeng and Zuo, Xiaochen and Yue, Yu and Dai, Weinan and Fan, Tiantian and Liu, Gaohong and Liu, Lingjun and others},
  journal={arXiv preprint arXiv:2503.14476},
  year={2025}
}

@article{shao2024deepseekmath,
  title={{DeepSeekMath}: Pushing the Limits of Mathematical Reasoning in Open Language Models},
  author={Shao, Zhihong and Wang, Peiyi and Zhu, Qihao and Xu, Runxin and Song, Junxiao and Bi, Xiao and Zhang, Haowei and Zhang, Mingchuan and Li, YK and Wu, Yang and others},
  journal={arXiv preprint arXiv:2402.03300},
  year={2024}
}

@misc{team2025qwen3,
  title={{Qwen3-VL}: Sharper Vision, Deeper Thought, Broader Action},
  author={{Qwen Team}},
  year={2025},
  howpublished={Qwen Blog. Accessed 2025-10-04},
}

@article{team2025gemma,
  title={{Gemma 3} Technical Report},
  author={Team, Gemma and Kamath, Aishwarya and Ferret, Johan and Pathak, Shreya and Vieillard, Nino and Merhej, Ramona and Perrin, Sarah and Matejovicova, Tatiana and Ram{\'e}, Alexandre and Rivi{\`e}re, Morgane and others},
  journal={arXiv preprint arXiv:2503.19786},
  year={2025}
}

@inproceedings{wang2025divide,
  title={Divide, Conquer and Combine: A Training-Free Framework for High-Resolution Image Perception in Multimodal Large Language Models},
  author={Wang, Wenbin and Ding, Liang and Zeng, Minyan and Zhou, Xiabin and Shen, Li and Luo, Yong and Yu, Wei and Tao, Dacheng},
  booktitle=AAAI,
  year={2025}
}

@article{li2023evaluating,
  title={Evaluating Object Hallucination in Large {Vision-Language Models}},
  author={Li, Yifan and Du, Yifan and Zhou, Kun and Wang, Jinpeng and Zhao, Wayne Xin and Wen, Ji-Rong},
  journal={arXiv preprint arXiv:2305.10355},
  year={2023}
}

@inproceedings{guan2024hallusionbench,
  title={{HallusionBench}: An Advanced Diagnostic Suite for Entangled Language Hallucination and Visual Illusion in Large {Vision-Language Models}},
  author={Guan, Tianrui and Liu, Fuxiao and Wu, Xiyang and Xian, Ruiqi and Li, Zongxia and Liu, Xiaoyu and Wang, Xijun and Chen, Lichang and Huang, Furong and Yacoob, Yaser and others},
  booktitle=CVPR,
  year={2024}
}

@inproceedings{yue2024mmmu,
  title={{MMMU}: A Massive Multi-Discipline Multimodal Understanding and Reasoning Benchmark for Expert {AGI}},
  author={Yue, Xiang and Ni, Yuansheng and Zhang, Kai and Zheng, Tianyu and Liu, Ruoqi and Zhang, Ge and Stevens, Samuel and Jiang, Dongfu and Ren, Weiming and Sun, Yuxuan and others},
  booktitle={Proceedings of the IEEE/CVF Conference on Computer Vision and Pattern Recognition},
  pages={9556--9567},
  year={2024}
}

@article{chen2024we,
  title={Are We on the Right Way for Evaluating Large Vision-Language Models?},
  author={Chen, Lin and Li, Jinsong and Dong, Xiaoyi and Zhang, Pan and Zang, Yuhang and Chen, Zehui and Duan, Haodong and Wang, Jiaqi and Qiao, Yu and Lin, Dahua and others},
  journal={Advances in Neural Information Processing Systems},
  volume={37},
  pages={27056--27087},
  year={2024}
}

@inproceedings{fu2024blink,
  title={{BLINK}: Multimodal Large Language Models Can See but Not Perceive},
  author={Fu, Xingyu and Hu, Yushi and Li, Bangzheng and Feng, Yu and Wang, Haoyu and Lin, Xudong and Roth, Dan and Smith, Noah A and Ma, Wei-Chiu and Krishna, Ranjay},
  booktitle={European Conference on Computer Vision},
  pages={148--166},
  year={2024},
  organization={Springer}
}

@inproceedings{kembhavi2016diagram,
  title={A Diagram Is Worth a Dozen Images},
  author={Kembhavi, Aniruddha and Salvato, Mike and Kolve, Eric and Seo, Minjoon and Hajishirzi, Hannaneh and Farhadi, Ali},
  booktitle={European Conference on Computer Vision},
  pages={235--251},
  year={2016},
  organization={Springer}
}

@misc{xai2024realworldqa,
  title={{RealWorldQA}: A Benchmark for Real-World Spatial Understanding},
  author={{xAI Team}},
  howpublished={\url{https://x.ai/blog/grok-1.5}},
  year={2024}
}
}

\section*{Appendix}

\section{AOT-SFT Dataset Construction Details}
\label{appendix:dataset_construction}

\begin{figure*}[htbp]
\centering
\includegraphics[width=\linewidth]{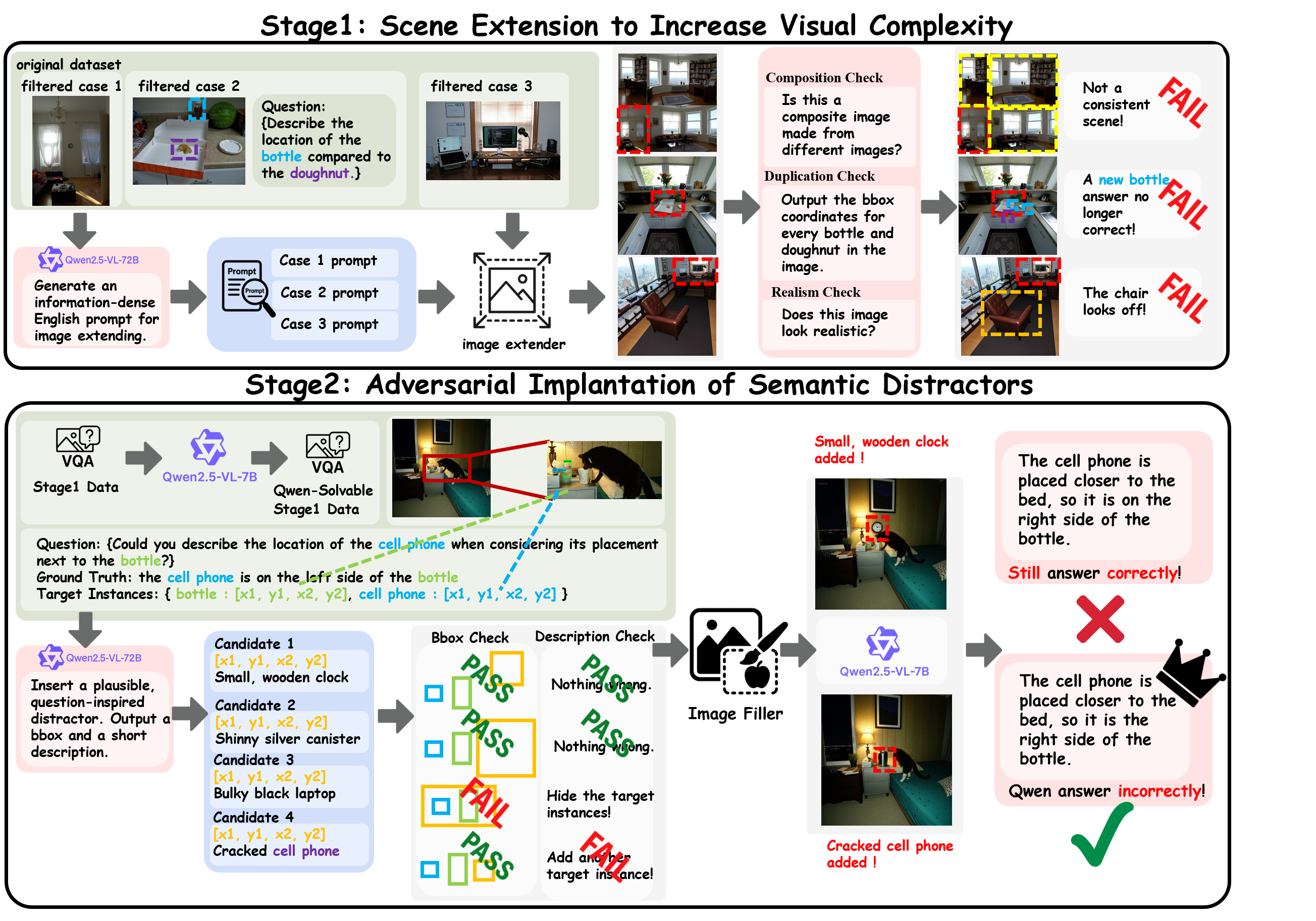}
\caption{An overview of our two-stage data generation pipeline for creating the AOT-SFT dataset. (Top) Stage 1 begins with an original image, which is then expanded by an image extender based on MLLM-generated prompts. The resulting image undergoes a three-part filtering process (Composition, Duplication, and Realism checks) to ensure quality and coherence. (Bottom) Stage 2 takes the validated images and uses an MLLM to propose semantic distractors. These proposals are filtered through Bbox and Description checks before being implanted by an image filler. The final adversarial image is validated for its effectiveness in fooling the defender model; only successful attacks are retained for the AOT-SFT dataset.}
\label{fig:sft_pipeline2}
\end{figure*}

This section provides a detailed breakdown of the implementation specifics for the two-stage data generation pipeline described in \Cref{sec:data_generation}, visually summarized in \Cref{fig:sft_pipeline2}. Our methodology relies heavily on programmatic queries to a powerful Multimodal Large Language Model (MLLM), specifically Qwen2.5-VL (72B), to perform various filtering and proposal generation tasks. The key settings, data preparation steps, and prompts used throughout the pipeline are detailed below to ensure full reproducibility.

\subsection{Stage 1: Scene Extension and Filtering Implementation}
\label{appendix:stage1_details}

As outlined in the main paper, the first stage involves expanding the canvas of images from the VStar dataset \cite{wu2023V*Bench} using an outpainting model guided by prompts from Qwen2.5-VL (72B). This process begins with a data augmentation step to prepare the images, followed by prompt generation and the three-part automated filtering.

\paragraph{Scene Extension Pre-processing.}
To create sufficient space for the outpainting model to generate new content, we first place each source image onto a larger, standardized canvas. This pre-processing is implemented as follows:

(1) \textit{Canvas Standardization}: A fixed-size canvas of \(1536 \times 1536\) pixels with a white background is created for every image.

(2) \textit{Image Resizing}: If an original image's width or height exceeds 1536 pixels, it is proportionally downscaled using a Lanczos resampling filter until it fits entirely within the canvas. The scaling ratio is recorded for subsequent coordinate adjustments.

(3) \textit{Stochastic Placement}: To increase the diversity of the generated scenes, the (potentially downscaled) image is placed onto the canvas according to a weighted random strategy: center placement (40\% probability), random placement (40\% probability), or corner placement (20\% probability).

(4) \textit{Mask Generation and Coordinate Transformation}: A corresponding binary mask is generated, where the original image area is black (masked) and the surrounding canvas is white (unmasked), indicating the region for outpainting. Crucially, all ground-truth bounding box coordinates from the original VStar annotation are transformed to align with the new canvas. This involves first applying the scaling ratio (if any) and then translating the coordinates based on the image's final position on the canvas.

\paragraph{Outpainting Prompt Generation.}
To guide the image extender model, we programmatically generate a detailed textual prompt for each padded image using Qwen2.5-VL (72B). The MLLM is instructed to act as an expert prompt engineer, describing a complete and coherent environment surrounding the visible portion of the image. A critical negative constraint is embedded in the prompt, explicitly forbidding the MLLM from mentioning the target objects already present in the original question. This is vital to preserve the ground-truth answer. The model is guided to produce a dense, comma-separated list of factual descriptions, prioritizing concrete nouns over abstract adjectives, with a target length of approximately 77 tokens. The full prompt template for this task is shown in \Cref{fig:prompt_outpainting}.

\paragraph{Hyperparameters and Technical Setup.}
The data processing pipeline was executed in a distributed, sharded manner to handle the large volume of images efficiently. Key parameters governing the pipeline's execution are summarized in \Cref{tab:pipeline_hyperparameters}. The core logic for filtering is implemented as a series of sequential checks, where an image is discarded immediately upon failing any single check. All MLLM queries incorporate a retry mechanism (up to 3 attempts) to handle transient API failures.

\begin{table}[htbp]
\centering
\small
\begin{tabular}{@{}ll@{}}
\toprule
Hyperparameter & Value \\
\midrule
Verification MLLM & Qwen2.5-VL (72B) \\
Canvas Size & \(1536 \times 1536\) pixels \\
\begin{tabular}[c]{@{}l@{}}Placement Weights\\(Center/Random/Corner)\end{tabular} & [0.4, 0.4, 0.2] \\
Composition Check Threshold & 0.5 (probability) \\
Realism Check Threshold & 0.5 (confidence) \\
Parallel Processing Workers & 16 per shard (filtering) \\
API Request Retries & 3 \\
\bottomrule
\end{tabular}
\caption{Key hyperparameters and settings for the Stage 1 data filtering pipeline. The thresholds represent the minimum confidence required from the MLLM to flag an image for removal.}
\label{tab:pipeline_hyperparameters}
\end{table}

\paragraph{Implementation of Filtering Checks.}
The three filtering checks—Composition, Duplication, and Realism—are implemented by sending specific prompts along with the candidate image to the Qwen2.5-VL (72B) model. The model's responses are then programmatically parsed to make a final decision. The exact prompts are provided to illustrate the precise instructions given to the model.

(1) \textit{Composition Check}: This check is designed to eliminate images that appear to be simple collages of unrelated scenes rather than a single, coherent environment. The prompt, shown in \Cref{fig:prompt_composition_check}, strictly defines what constitutes a simple stitched collage and instructs the model to provide a `Yes' or `No' answer. We calculate the probability of the `Yes' token and discard the image if this probability exceeds a threshold of \(0.5\).

(2) \textit{Duplication Check}: This check ensures that the outpainting process did not introduce additional instances of the target objects mentioned in the question. We query the MLLM to detect all instances of a specific target object class within the image. The prompt, detailed in \Cref{fig:prompt_duplication_check}, emphasizes high precision and avoidance of false positives. The model is asked to return a JSON object containing the bounding boxes of all detected instances. The image is discarded if the number of returned bounding boxes is greater than the number of ground-truth instances in the original image.

(3) \textit{Realism Check}: This check filters out images containing obvious AI-generated artifacts or logical inconsistencies. The prompt, shown in \Cref{fig:prompt_realism_check}, instructs the model to act as a conservative digital forensics expert, only flagging images with undeniable evidence of being synthetic. The model must begin its response with either `Real Photo' or `AI-Generated'. Any image classified as `AI-Generated' is discarded.

\begin{figure*}[htbp]
    \centering
    \begin{minipage}{0.9\linewidth}
    \small
    \texttt{You are an expert prompt engineer for advanced AI image generation models. You will be given a partial image on a white canvas. Your mission is to write a single, richly detailed, and factual English prompt to outpaint the surrounding area, creating a complete, coherent, and visually interesting scene.}
    
    \vspace{1ex}
    \texttt{CRITICAL INSTRUCTION}
    
    \texttt{The visible photograph already contains: **\{exclusion\_list\}**.}
    
    \texttt{**\underline{DO NOT mention these objects in your output prompt.}** Your goal is to describe the environment *around* them.}
    
    \vspace{1ex}
    \texttt{PROMPT STYLE \& CONTENT GUIDELINES}
    
    \texttt{1. Maximize Detail (Token Budget): Your output prompt should be dense with information, aiming to utilize a budget of **around 77 tokens**. Do not be brief; your goal is to describe a full scene.}
    
    \texttt{2. Describe a Complete Environment: Instead of just one or two items, describe a whole setting. **Add more distinct objects and environmental details.** Think about what would be in the foreground, mid-ground, and background. What's on the walls? What's on the floor? Is there a view outside a window?}
    
    \texttt{3. Prioritize Nouns over Adjectives: Build the scene with a variety of concrete objects, structures, and scenery (e.g., `a mahogany desk, a leather armchair, a large bay window overlooking a city skyline, bookshelves filled with books, a patterned rug on a hardwood floor'). This is more effective than using many adjectives on few items.}
    
    \texttt{4. Strict Comma-Separated Format: Continue to structure your output as a series of descriptive phrases separated by commas.}
    
    \texttt{5. Factual \& Unambiguous: Describe *what is there*, not *how it feels*. Avoid abstract or emotional terms like `cozy', `beautiful', or `serene'.}
    \end{minipage}
    \caption{The prompt used for generating outpainting descriptions in Stage 1. The placeholder \texttt{\{exclusion\_list\}} is populated with the names of the target objects from the question.}
    \label{fig:prompt_outpainting}
\end{figure*}

\begin{figure*}[htbp]
    \centering
    \begin{minipage}{0.9\linewidth}
    \small
    \texttt{Your task is to identify if this image is a *simple stitched collage*. A simple stitched collage is strictly defined as two or more completely separate and unrelated photographs joined together at their edges to fill the canvas (e.g., side-by-side, top-and-bottom grids). The key characteristic is the lack of a single, unifying `host' scene.}
    
    \vspace{1ex}
    \texttt{CRITICAL RULE: If the image depicts a single, continuous `host' scene that itself *contains* another image or view, it is **NOT** a simple stitched collage. Examples of a valid single scene include: - A room with a picture frame on the wall. - A view through a window or a reflection in a mirror. - An artistic frame-within-a-frame or photo-in-photo composition. - A scene with a minor visual artifact, like a thin line, that doesn't break it into two distinct, unrelated environments.}
    
    \vspace{1ex}
    \texttt{Based on this strict definition, is this a simple stitched collage? Your answer MUST begin with `Yes' or `No'.}
    \end{minipage}
    \caption{The prompt used for the Composition Check. The model's log probabilities for the first token are used to determine the confidence in the `Yes' classification.}
    \label{fig:prompt_composition_check}
\end{figure*}

\begin{figure*}[htbp]
    \centering
    \begin{minipage}{0.9\linewidth}
    \small
    \texttt{You are a highly precise visual detection expert. Your primary goal is to avoid false positives. Your task is to identify and provide bounding boxes for *only* the instances of the object `\{object\_name\}' in the image.}

    \vspace{1ex}
    \texttt{CRITICAL RULES:}
    
    \texttt{1. Strict Identification: Be extremely strict. If an object is merely *similar* to a `\{object\_name\}' but is technically a different category (e.g., a `truck' is NOT a `car'; a `sofa' is NOT a `chair'), you MUST NOT include it.}
    
    \texttt{2. Full Definition: Pay close attention to the complete term. For example, a `potted plant' must be a plant physically situated *inside a pot*. A plant hanging on a wall or growing in the ground is not a `potted plant'.}
    
    \texttt{3. High Confidence Only: If you are not absolutely certain that an object perfectly matches the description, DO NOT return a bounding box for it. It is better to return an empty list than an incorrect box.}

    \vspace{1ex}
    \texttt{Respond with a JSON object containing a single key detections. The value should be a list of bounding boxes. Each item in the list must be a dictionary with a box key, and the value should be a list of four integers: [x, y, width, height]. If no objects are found, return an empty list: \{detections: []\}.}
    \end{minipage}
    \caption{The prompt used for the Duplication Check. The placeholder \texttt{\{object\_name\}} is replaced with the name of the target instance from the question.}
    \label{fig:prompt_duplication_check}
\end{figure*}

\begin{figure*}[htbp]
    \centering
    \begin{minipage}{0.9\linewidth}
    \small
    \texttt{You are a meticulous digital image forensics expert. Your task is to determine if the image is a real-world photograph or a synthetic (AI-generated) image.}

    \vspace{1ex}
    \texttt{CRITICAL CONSERVATIVE RULE: Your primary directive is to be conservative. You should only classify an image as `AI-Generated' if you find clear, undeniable evidence of digital artifacts commonly associated with AI image synthesis. If there is any ambiguity, you MUST default to classifying it as a `Real Photo'.}

    \vspace{1ex}
    \texttt{Look for these specific, unambiguous artifacts:}
    
    \texttt{1. Anatomical Errors: Malformed hands with incorrect numbers of fingers, distorted limbs, unnatural body poses.}
    
    \texttt{2. Text and Symbol Corruption: Garbled, nonsensical, or partially formed text and symbols.}
    
    \texttt{3. Logical Inconsistencies: Objects blending or melting into each other unnaturally, impossible physics, inconsistent lighting or shadows that are physically impossible.}
    
    \texttt{4. Unnatural Textures: Surfaces that look overly smooth, plastic-like, or have a characteristic waxy AI sheen.}

    \vspace{1ex}
    \texttt{RESPONSE FORMAT:}
    
    \texttt{- Your response MUST begin with either `Real Photo' or `AI-Generated'.}
    
    \texttt{- If you classify it as `AI-Generated', you MUST provide a brief, one-sentence explanation after a colon. Example: `AI-Generated: The text on the sign is garbled and unreadable.'}

    \vspace{1ex}
    \texttt{Analyze the image and provide your verdict based on these strict, conservative criteria.}
    \end{minipage}
    \caption{The prompt used for the Realism Check. This prompt encourages a conservative approach to minimize the false rejection of plausible-looking generated scenes.}
    \label{fig:prompt_realism_check}
\end{figure*}

\subsection{Stage 2: Adversarial Implantation Implementation}
\label{appendix:stage2_details}

In the second stage, we focus on implanting semantic distractors into the images that successfully passed all Stage 1 filters. This stage involves generating distractor proposals, performing integrity checks, implanting the object, and finally validating its adversarial efficacy.

\paragraph{Clean Candidate Pool Formation.}
Before generating distractors, we first filter the extended images \(I'\) to form a clean candidate pool, \(D_{clean}\). We achieve this by feeding each pair \((I', Q)\) to our initial defender model, \(M_{def}^{(0)}\). Only those samples for which \(M_{def}^{(0)}\) produces the correct answer are included in \(D_{clean}\). This critical step ensures that our subsequent adversarial attacks are generated against scenarios where the model is initially competent, thereby measuring true vulnerability to distractors rather than existing model failures.

\paragraph{Distractor Proposal Generation.}
For each sample in \(D_{clean}\), we use Qwen2.5-VL (72B) to propose a set of potential distractors. The MLLM is prompted to suggest a plausible object that could mislead a model answering the given question. The prompt, shown in \Cref{fig:prompt_distractor_proposal}, explicitly instructs the model to provide both a bounding box \([x_1, y_1, x_2, y_2]\) and a short English description, and to avoid obscuring evidence for the correct answer. The raw text response from the MLLM is parsed using a regular expression to extract these two components. To ensure diversity and prevent redundant processing, a unique fingerprint is generated for each valid proposal based on the image path, rounded bounding box coordinates, and description text. Duplicate proposals are discarded.

\paragraph{Integrity Checks for Distractor Proposals.}
Each successfully parsed proposal undergoes a series of programmatic checks before being rendered by the inpainting model:

(1) \textit{Bbox Validity Check}: The parsed bounding box undergoes geometric validation. It must be well-formed (\(x_1 < x_2, y_1 < y_2\)), lie entirely within the image boundaries, and exceed a minimum area threshold of 100 square pixels to prevent trivial or point-like suggestions.

(2) \textit{Bbox Overlap Check}: This is a geometric check to prevent the distractor from occluding relevant objects. We compute the Intersection over Union (IoU) between the proposed bounding box \(B\) and all ground-truth bounding boxes of the target instances in the image. If the IoU for any ground-truth box is greater than a threshold of \(0.0\), the proposal is rejected. This strict zero-tolerance policy ensures that the distractor does not physically interfere with the objects central to the question.

(3) \textit{Description Check}: This is a string-matching check to prevent the generation of another instance of a target object. We perform a case-insensitive keyword search within the object description prompt \(P_{obj}\) after removing common English stopwords. If any of the keywords corresponding to the target instance names are found in \(P_{obj}\), the proposal is rejected. This maintains the logical premise of the original question (e.g., preventing the addition of a second apple when the question is How many apples are there?).

\paragraph{Image Generation Backend.}
Both the scene extension in Stage 1 and the distractor implantation in Stage 2 are executed using a powerful diffusion-based inpainting model, specifically the OneReward \cite{gong2025onereward}. The model takes the padded image, the corresponding binary mask, and the MLLM-generated textual prompt as input to generate the final high-resolution image. 

\paragraph{Final Efficacy Validation.}
Proposals that pass all integrity checks are used to generate a candidate adversarial image \(I_{adv}\) via inpainting. This candidate is then validated for its adversarial effectiveness. The pair \((I_{adv}, Q)\) is passed to the same initial defender model, \(M_{def}^{(0)}\). If the model's prediction is now incorrect, the distractor is confirmed to be effective. Only these successful attack triplets, \((I', Q, I_{adv})\), are added to the final AOT-SFT dataset. If the model still answers correctly, the candidate is discarded as an ineffective attack.

\begin{figure*}[htbp]
    \centering
    \begin{minipage}{0.9\linewidth}
    \small
    \texttt{<image>Add a plausible distractor object, inspired by the question, to mislead a model answering: \{question\}.}
    
    \vspace{1ex}
    \texttt{The object's description must not contain key nouns from the question. Your bounding box [x1, y1, x2, y2] must not obscure evidence for the correct answer.}
    
    \vspace{1ex}
    \texttt{Respond in a single line: [x1, y1, x2, y2] | A short English description.}
    \end{minipage}
    \caption{The prompt used for generating distractor proposals in Stage 2. The placeholder \texttt{\{question\}} is populated with the question associated with the image.}
    \label{fig:prompt_distractor_proposal}
\end{figure*}

\section{Iterative Co-evolution Implementation Details}
\label{appendix:co_evolution_details}

This section provides a detailed description of the implementation for the iterative co-evolutionary framework introduced in \Cref{sec:co_evolution}. We elaborate on the specific models, algorithms, reward function implementations, and hyperparameters used to train both the attacker and the defender models.

\subsection{Attacker Evolution Implementation}
\label{appendix:attacker_evolution_impl}

The attacker model, \(M_{atk}\), is an instance of the Qwen-Image-Edit model. Its policy is optimized using the Flow-GRPO algorithm \cite{liu2025flow}. The following paragraphs detail the technical implementation of its reward function and the subsequent process for curating the adversarial training set for the defender.

\paragraph{Reward Function Implementation.}
The composite reward function \(R_{atk}\), formally defined in \Cref{eq:reward_atk}, is implemented through a sequence of programmatic checks. The specific values and logic are derived from our codebase to ensure reproducibility.

(1) \textit{Semantic Integrity Check}: The localized SSIM check is the first and most critical step. For each generated image \(I_{adv}\), we compare image patches corresponding to the ground-truth bounding boxes of critical objects against the original image \(I'\). Based on our empirical findings, the SSIM threshold is set to \(\tau_{ssim} = 0.3\). If the SSIM score for any single critical region falls below this threshold, the process is immediately terminated for that sample, and a reward of \(0\) is assigned. This strict, per-instance check is crucial for preventing the attacker from learning trivial solutions, such as erasing or fundamentally altering key objects.

(2) \textit{Adversarial Efficacy Check}: For candidates that pass the SSIM check, we evaluate their ability to deceive the current defender, \(M_{def}^{(i-1)}\). As described in the main text, an attack is only considered successful if it consistently fools the defender. This is implemented by performing two consecutive inference passes with the defender model using deterministic decoding (i.e., temperature set to 0). If the defender answers incorrectly in both attempts, the full reward of \(1.0\) is granted. If the defender answers correctly in at least one of the two attempts, the attack is considered valid but ineffective, earning a base reward of \(0.2\). This small incentive encourages the attacker to explore the space of valid, non-trivial image edits.

\paragraph{Adversarial Set Curation via Stochastic Evaluation.}
The process of curating the training set \(D_{adv}^{(i)}\) for the defender involves a more nuanced evaluation to identify challenging yet learnable examples. This filtering mechanism is implemented as follows:

(1) \textit{Multiple Inference Trials}: For each adversarial candidate \(I_{adv}\) generated by the updated attacker \(M_{atk}^{(i)}\), we run \(10\) separate inference trials against the frozen, previous-generation defender \(M_{def}^{(i-1)}\).

(2) \textit{Stochastic Sampling}: Unlike the deterministic check used in the attacker's reward function, these trials are performed using stochastic sampling with the temperature set to \(1.0\). This approach provides a more robust measure of the defender's confidence and stability by exploring different generation paths.

(3) \textit{Difficulty Windowing}: The core of the filtering logic lies in a difficulty window. We count the number of times the defender produces the correct answer across the \(10\) trials. An adversarial example is added to the training set \(D_{adv}^{(i)}\) only if this count falls within the inclusive range of \([3, 7]\). This strategy effectively discards examples that are too easy (correct answers \(> 70\%\)) or too hard (correct answers \(< 30\%\))), focusing the defender's training on samples that lie closest to its decision boundary.

\subsection{Defender Enhancement Implementation}
\label{appendix:defender_enhancement_impl}

The defender model, \(M_{def}\), is the LVLM we aim to fortify. Its training is driven by the DAPO algorithm \cite{yu2025dapo, shao2024deepseekmath}, using the adversarial dataset \(D_{adv}^{(i)}\) curated in the previous step.

\paragraph{Reward Function Implementation.}
The defender's reward function, \(R_{def}\), as defined in \Cref{eq:reward_def}, is designed to promote both correctness and adherence to specific output formatting, which is crucial for reliable instruction-following behavior in multiple-choice VQA tasks.

(1) \textit{Robust Answer Extraction}: To determine correctness, we first need to robustly parse the answer from the model's free-form text output. We implement a cascaded regular expression strategy. The primary pattern targets the canonical LaTeX format, `\(\texttt{\string\boxed{LETTER}}\)`, where `LETTER` is a capital letter (e.g., `A', `B'). If this pattern is not found, a series of fallback patterns are used to find other common formats, such as `\(\texttt{boxed\{A\}}\)`, `\(\texttt{The final answer is: A}\)`, or `\(\texttt{(A)}\)`. An answer is considered successfully extracted if any of these patterns match.

(2) \textit{Scoring Logic}: If the extracted answer matches the ground-truth answer, a base reward of \(0.8\) is awarded. An additional format bonus of \(0.2\) is given only if the answer was extracted using the primary, perfect `\(\texttt{\string\boxed{LETTER}}\)` pattern. This two-tiered reward structure strongly incentivizes the model to not only be correct but also to present its final answer in the most clear and unambiguous format, while not overly penalizing correct answers that are formatted differently. If the extracted answer is incorrect or if no answer can be extracted, the reward is \(0\).

\paragraph{Hyperparameters.}
The key hyperparameters governing the iterative co-evolution process are summarized in \Cref{tab:coevolution_hyperparameters}. These values were determined through preliminary experiments to balance the learning stability of both the attacker and the defender.

\begin{table}[htbp]
\centering
\small
\begin{tabular}{@{}ll@{}}
\toprule
Hyperparameter & Value \\
\midrule
\multicolumn{2}{@{}l}{\textit{Attacker Evolution} (\(M_{atk}\))} \\
Algorithm & Flow-GRPO \\
SSIM Threshold (\(\tau_{ssim}\)) & 0.3 \\
Base Reward (Valid Edit) & 0.2 \\
Full Reward (Successful Attack) & 1.0 \\
\begin{tabular}[c]{@{}l@{}}Filtering VQA Attempts\\(for \(D_{adv}^{(i)}\) curation)\end{tabular} & 10 \\
Filtering VQA Temperature & 1.0 \\
Filtering Min Correct Answers & 3 \\
Filtering Max Correct Answers & 7 \\
\midrule
\multicolumn{2}{@{}l}{\textit{Defender Enhancement} (\(M_{def}\))} \\
Algorithm & DAPO \\
Accuracy Reward & 0.8 \\
Format Bonus Reward & 0.2 \\
\bottomrule
\end{tabular}
\caption{Key hyperparameters for the iterative co-evolution training framework. These settings control the reward signals and filtering criteria for both the attacker and defender models.}
\label{tab:coevolution_hyperparameters}
\end{table}

\section{Additional Experimental Results}
\label{appendix:additional_results}

In this section, we provide a comprehensive breakdown of the performance metrics across all evaluated benchmarks. We report the detailed sub-category scores to offer a granular view of the improvements achieved by our co-evolutionary framework compared to the baselines.

\subsection{Detailed Results on Diagram and Visual Perception}

\Cref{tab:ai2d_detailed} presents the fine-grained results on the AI2D benchmark. Our method demonstrates consistent improvements across diverse diagram types, particularly in complex categories such as ``Food Chains/Webs'' and ``Life Cycles,'' where understanding the structural relationships is key.

\Cref{tab:blink_detailed} details the performance on the BLINK benchmark, which evaluates multifaceted visual perception. Notably, our model achieves substantial gains in the ``IQ Test'' and ``Relative Depth'' categories in the final iteration, indicating an enhanced ability to reason about abstract patterns and spatial geometry.

\begin{table*}[h]
\centering
\caption{Detailed performance breakdown on the \textbf{AI2D} benchmark. We report accuracy (\%) for each diagram type.}
\label{tab:ai2d_detailed}
\resizebox{\textwidth}{!}{%
\begin{tabular}{@{}l c cccccccccccccccc@{}}
\toprule
\textbf{Method} & \textbf{Overall} & \textbf{\makecell{Atom\\Struct.}} & \textbf{Eclipses} & \textbf{\makecell{Faults\\Quakes}} & \textbf{\makecell{Food\\Chains}} & \textbf{\makecell{Life\\Cycles}} & \textbf{\makecell{Moon\\Phase}} & \textbf{\makecell{Parts\\of A}} & \textbf{\makecell{Parts\\of Earth}} & \textbf{\makecell{Photo.\\Resp.}} & \textbf{\makecell{Rock\\Cycle}} & \textbf{\makecell{Rock\\Strata}} & \textbf{\makecell{Solar\\System}} & \textbf{\makecell{Types\\of}} & \textbf{Volcano} & \textbf{\makecell{Water\\Cycle}} \\
\midrule
Base (\(M_{def}^{(0)}\)) & 80.96 & 75.00 & 87.10 & 53.57 & 92.51 & 75.06 & 68.23 & 78.44 & 84.62 & 67.09 & 62.69 & 75.61 & 88.89 & 73.18 & 100.0 & 54.55 \\
+ Clean Data & 80.73 & 75.00 & 87.10 & 53.57 & 92.25 & 74.34 & 66.43 & 79.06 & 84.62 & 68.35 & 62.69 & 75.61 & 88.89 & 73.18 & 100.0 & 54.55 \\
Defender Iter. 1 & 81.09 & 75.00 & 87.10 & 53.57 & 92.25 & 76.02 & 68.59 & 78.64 & 84.62 & 68.35 & 61.19 & 75.61 & 88.89 & 73.47 & 100.0 & 54.55 \\
Defender Iter. 2 & 81.25 & 75.00 & 87.10 & 53.57 & 92.77 & 76.50 & 66.79 & 78.64 & 84.62 & 69.62 & 61.19 & 78.05 & 88.89 & 73.47 & 100.0 & 54.55 \\
Defender Iter. 3 & \textbf{81.35} & 75.00 & 87.10 & 53.57 & \textbf{92.77} & \textbf{76.26} & 66.79 & 78.85 & 84.62 & \textbf{70.89} & \textbf{64.18} & \textbf{78.05} & 88.89 & 73.47 & 100.0 & 54.55 \\
\bottomrule
\end{tabular}%
}
\end{table*}

\begin{table*}[h]
\centering
\caption{Detailed performance breakdown on the \textbf{BLINK} benchmark. We report accuracy (\%) across all visual tasks.}
\label{tab:blink_detailed}
\resizebox{\textwidth}{!}{%
\begin{tabular}{@{}l c ccccccccccccccc@{}}
\toprule
\textbf{Method} & \textbf{Overall} & \textbf{\makecell{Art\\Style}} & \textbf{Count} & \textbf{\makecell{Forensic\\Detect}} & \textbf{\makecell{Func.\\Corr.}} & \textbf{\makecell{IQ\\Test}} & \textbf{Jigsaw} & \textbf{\makecell{Multi-\\view}} & \textbf{\makecell{Obj.\\Loc.}} & \textbf{\makecell{Rel.\\Depth}} & \textbf{\makecell{Rel.\\Reflect}} & \textbf{\makecell{Sem.\\Corr.}} & \textbf{\makecell{Spatial\\Rel.}} & \textbf{\makecell{Vis.\\Corr.}} & \textbf{\makecell{Vis.\\Sim.}} \\
\midrule
Base (\(M_{def}^{(0)}\)) & 54.08 & 70.09 & 70.00 & 48.48 & 26.92 & 0.67 & 59.33 & 54.14 & 51.64 & 83.06 & 38.81 & 33.81 & 88.81 & 53.49 & 86.67 \\
+ Clean Data & 54.55 & 70.94 & 71.67 & 49.24 & 24.62 & 2.00 & 63.33 & 51.88 & 53.28 & 79.84 & 36.57 & 35.97 & 86.01 & 58.72 & 86.67 \\
Defender Iter. 1 & 54.55 & 69.23 & 70.83 & 47.73 & 24.62 & 1.33 & 62.00 & 54.89 & 52.46 & 80.65 & 38.81 & 35.25 & 88.11 & 58.14 & 86.67 \\
Defender Iter. 2 & 54.76 & 70.94 & 72.50 & 50.00 & 23.08 & 5.33 & 62.00 & 52.63 & 53.28 & 80.65 & 38.81 & 33.81 & 86.01 & 58.72 & 85.93 \\
Defender Iter. 3 & \textbf{55.92} & 70.09 & \textbf{72.50} & 49.24 & 26.15 & \textbf{10.00} & \textbf{70.00} & 54.89 & 53.28 & 82.26 & 38.06 & 33.81 & 86.01 & 55.81 & \textbf{87.41} \\
\bottomrule
\end{tabular}%
}
\end{table*}

\subsection{Real-World Application Performance}

We further investigate the model's capability in real-world scenarios using the MME-RealWorld-Lite benchmark. We separate the analysis into Reasoning tasks (\Cref{tab:mme_reasoning}) and Perception tasks (\Cref{tab:mme_perception}).

The results highlight the efficacy of our approach in handling complex, noisy real-world data. In the Perception tasks, our final model achieves a notable \textbf{55.35\%} Overall score, significantly outperforming the base model (48.85\%) and the clean data baseline (53.64\%). Improvements are particularly strong in ``Remote Sensing'' and ``Autonomous Driving,'' where robustness to visual distractors and noise is critical.

\begin{table*}[h]
\centering
\caption{Detailed performance on \textbf{MME-RealWorld-Lite (Reasoning Tasks)}. We report the overall Reasoning score and breakdowns for specific domains.}
\label{tab:mme_reasoning}
\begin{tabular}{@{}l c ccccc@{}}
\toprule
\textbf{Method} & \textbf{\makecell{Overall\\Reasoning}} & \textbf{Monitoring} & \textbf{\makecell{Autonomous\\Driving}} & \textbf{\makecell{OCR\\Context}} & \textbf{\makecell{Diagram\\\& Table}} & \textbf{\makecell{Remote\\Sensing}} \\
\midrule
Base (\(M_{def}^{(0)}\)) & 37.47 & 27.33 & 25.50 & 73.00 & 65.00 & 0.00 \\
+ Clean Data & 42.53 & 36.67 & 30.75 & 75.00 & 66.00 & 0.00 \\
Defender Iter. 1 & 42.13 & 34.00 & 30.25 & 73.00 & 71.00 & 0.00 \\
Defender Iter. 2 & 41.33 & 34.00 & 29.50 & 74.00 & 67.00 & 0.00 \\
Defender Iter. 3 & \textbf{44.00} & \textbf{40.67} & \textbf{31.75} & 73.00 & \textbf{69.00} & 0.00 \\
\bottomrule
\end{tabular}
\end{table*}

\begin{table*}[h]
\centering
\caption{Detailed performance on \textbf{MME-RealWorld-Lite (Perception Tasks)}. We report the overall Perception score and breakdowns for specific domains.}
\label{tab:mme_perception}
\begin{tabular}{@{}l c ccccc@{}}
\toprule
\textbf{Method} & \textbf{\makecell{Overall\\Perception}} & \textbf{Monitoring} & \textbf{\makecell{Autonomous\\Driving}} & \textbf{\makecell{OCR\\Context}} & \textbf{\makecell{Diagram\\\& Table}} & \textbf{\makecell{Remote\\Sensing}} \\
\midrule
Base (\(M_{def}^{(0)}\)) & 48.85 & 28.84 & 30.57 & 90.40 & 82.00 & 42.67 \\
+ Clean Data & 53.64 & 38.24 & 32.86 & \textbf{91.20} & 88.00 & 49.33 \\
Defender Iter. 1 & 53.29 & 35.42 & 35.14 & \textbf{91.20} & 88.00 & 47.33 \\
Defender Iter. 2 & 53.21 & 36.36 & 33.71 & 90.40 & 88.00 & 49.33 \\
Defender Iter. 3 & \textbf{55.35} & \textbf{39.81} & \textbf{36.57} & \textbf{91.20} & \textbf{89.00} & \textbf{50.00} \\
\bottomrule
\end{tabular}
\end{table*}

\subsection{Results on General Multimodal Benchmarks}

Finally, \Cref{tab:general_benchmarks_combined} consolidates the results across a broad spectrum of multimodal tasks, including the MMMU benchmark for collegiate-level knowledge, MMStar, SEEDBench2-Plus, RealWorldQA, and HRBench-8K.

For MMMU, we report the overall accuracy on both the validation and development splits. Despite the high difficulty of this benchmark, our method shows a positive trend, with the final defender achieving the highest accuracy on the development set (\textbf{25.33\%}) and competitive performance on the validation set.

On other benchmarks, our iterative approach yields steady improvements across both coarse and fine-grained metrics. Notably, on HRBench-8K, the ``Single'' accuracy improves from 78.75\% (Base) to 85.00\% (Iter. 3), demonstrating that our training curriculum significantly enhances fundamental visual acuity. Similarly, we observe consistent gains on SEEDBench2-Plus (reaching \textbf{70.05\%}) and RealWorldQA (\textbf{70.07\%}), confirming the generalizability of our method.

\begin{table*}[h]
\centering
\caption{Consolidated performance on general multimodal benchmarks. We report the Overall accuracy for \textbf{MMMU} (Val/Dev), \textbf{RealWorldQA}, and detailed breakdowns for \textbf{MMStar}, \textbf{SEEDBench2-Plus}, and \textbf{HRBench-8K}. The best results are highlighted in \textbf{bold}.}
\label{tab:general_benchmarks_combined}
\setlength{\tabcolsep}{2.5pt} 
\begin{tabular}{@{}l cc ccc cccc c ccc@{}}
\toprule
\multirow{3}{*}{\textbf{Method}} & \multicolumn{2}{c}{\textbf{MMMU}} & \multicolumn{3}{c}{\textbf{MMStar}} & \multicolumn{4}{c}{\textbf{SEEDBench2-Plus}} & \textbf{RWQA} & \multicolumn{3}{c}{\textbf{HRBench-8K (Avg)}} \\
\cmidrule(lr){2-3} \cmidrule(lr){4-6} \cmidrule(lr){7-10} \cmidrule(lr){11-11} \cmidrule(lr){12-14}
& \textbf{Val} & \textbf{Dev} & \textbf{Overall} & \textbf{Coarse} & \textbf{Fine} & \textbf{Overall} & \textbf{Chart} & \textbf{Map} & \textbf{Web} & \textbf{Overall} & \textbf{All} & \textbf{Cross} & \textbf{Single} \\
\midrule
Base (\(M_{def}^{(0)}\)) & 18.56 & 20.67 & 60.33 & 71.60 & 60.40 & 69.52 & 67.41 & 60.10 & \textbf{83.64} & 67.71 & 64.88 & 51.00 & 78.75 \\
+ Clean Data & 22.56 & 21.33 & 60.93 & \textbf{72.40} & 61.20 & 69.78 & 67.53 & \textbf{61.09} & 83.18 & 69.28 & 70.62 & 57.00 & 84.25 \\
Defender Iter. 1 & 21.56 & 20.00 & 61.27 & 71.20 & 59.60 & 69.78 & 67.90 & 60.72 & 83.18 & 69.28 & 69.38 & 55.25 & 83.50 \\
Defender Iter. 2 & \textbf{24.67} & 23.33 & 61.40 & 72.00 & 60.40 & 69.65 & 67.65 & 60.59 & 83.18 & 69.28 & 69.88 & 56.25 & 83.50 \\
Defender Iter. 3 & 23.67 & \textbf{25.33} & \textbf{61.53} & \textbf{72.40} & \textbf{61.20} & \textbf{70.05} & \textbf{68.40} & 60.97 & 83.18 & \textbf{70.07} & \textbf{71.50} & \textbf{58.00} & \textbf{85.00} \\
\bottomrule
\end{tabular}
\end{table*}

\section{Additional Qualitative Visualization}
\label{appendix:qualitative_viz}

\begin{figure*}[p]
    \centering
    \begin{subfigure}[b]{0.49\textwidth}
        \centering
        \includegraphics[width=\linewidth]{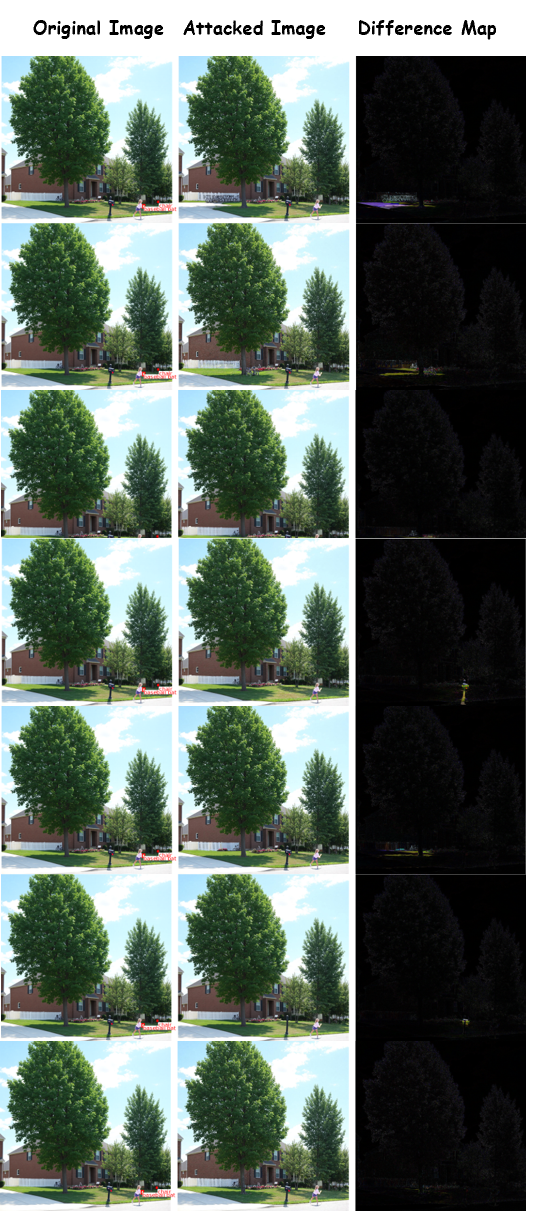}
        \caption{Scenario A: Suburban Scene}
        \label{fig:supp_attack_vis_1}
    \end{subfigure}
    \hfill 
    \begin{subfigure}[b]{0.49\textwidth}
        \centering
        \includegraphics[width=\linewidth]{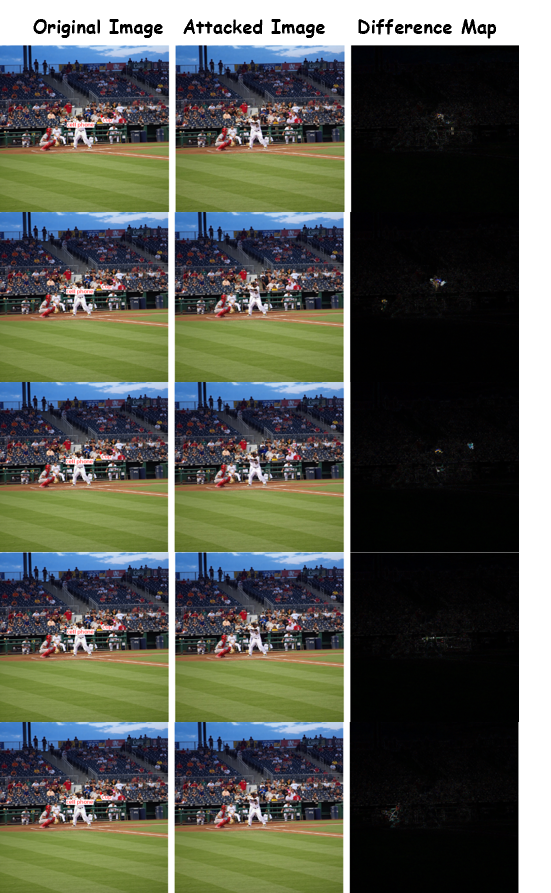}
        \caption{Scenario B: Sports Field Scene}
        \label{fig:supp_attack_vis_2}
    \end{subfigure}
    
    \caption{Extended qualitative examples of adversarial attacks generated by \(M_{atk}\). We visualize two distinct scenarios: \textbf{(a)} a suburban environment and \textbf{(b)} a sports field. In both cases, the columns display (from left to right): the original clean image, the attacked image with adversarial perturbations, and the contrast-enhanced difference map. The visualizations confirm that our attacker consistently learns to inject localized semantic features—such as specific object patches or textural changes—rather than relying on imperceptible global noise, demonstrating robust generalization across different image contexts.}
    \label{fig:supp_attack_vis_combined}
\end{figure*}

To further illustrate the capabilities of our evolved attacker, we provide additional qualitative examples of generated adversarial images in \Cref{fig:supp_attack_vis_combined}. These examples demonstrate the attacker's ability to manipulate visual semantics effectively while maintaining high image quality across different scenes.

As shown in the difference maps, the perturbations generated by our method are highly localized and sparse. Unlike traditional adversarial attacks that often rely on global high-frequency noise, our attacker, optimized through the proposed co-evolutionary framework, tends to focus on semantic regions. This includes synthesizing distinct objects or altering specific textures, which are captured by the high-contrast regions in the difference maps. These modifications are designed to be potent enough to mislead the defender model while remaining subtle or plausible to the human observer.

\begin{figure*}[p]
\centering
\includegraphics[width=\textwidth, height=0.95\textheight, keepaspectratio]{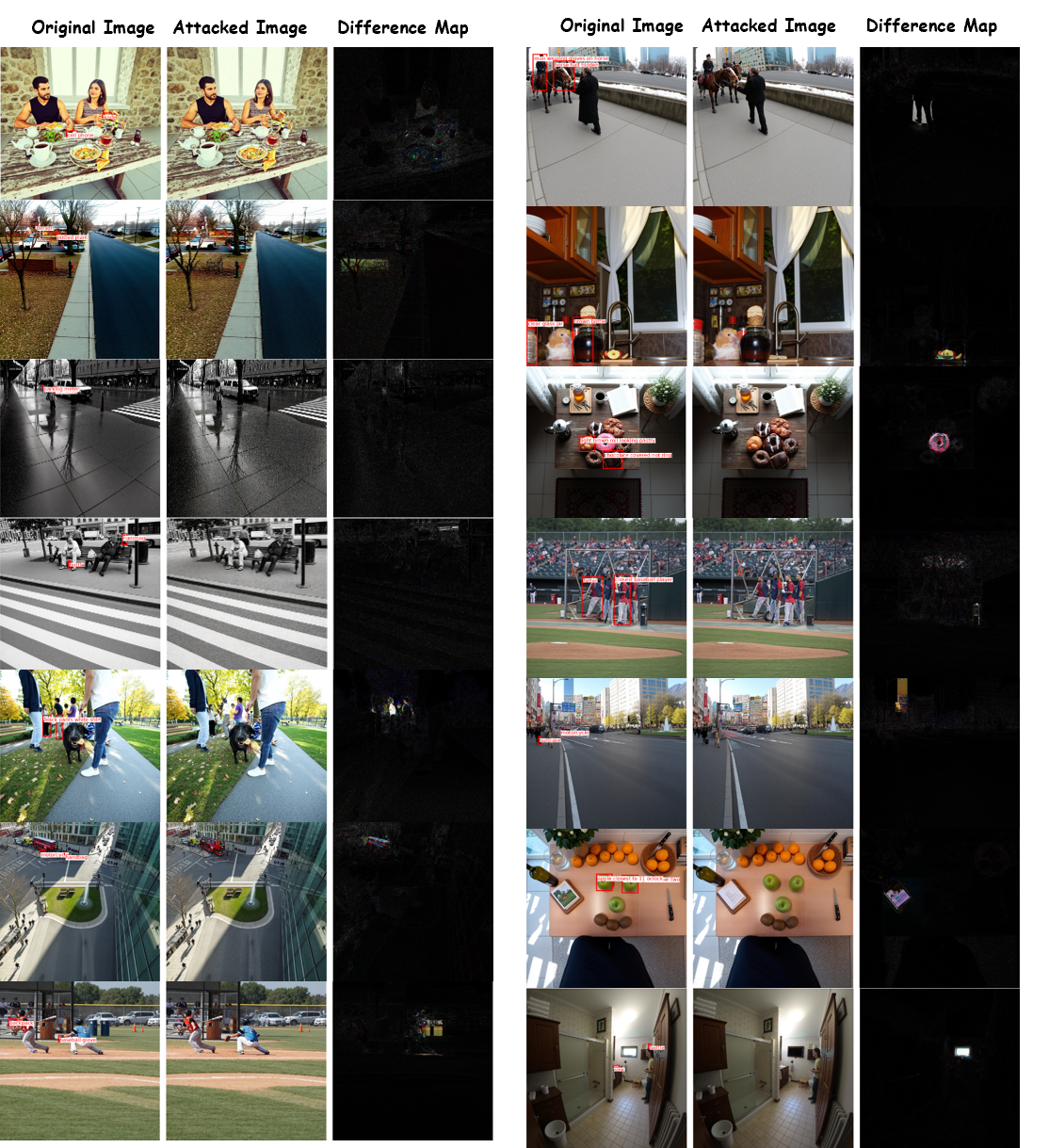}
\caption{Demonstration of the attacker's generalization capability across diverse visual scenes. Unlike \Cref{fig:supp_attack_vis_combined}, which showed multiple attack variations for fixed inputs, this figure displays a random selection of successful attacks on \textbf{distinct, unrelated images}. The samples cover a broad spectrum of domains, including complex indoor settings, outdoor street views, and dynamic sports scenes. In each instance, the attacker successfully identifies a semantic vulnerability and injects a localized perturbation (visible in the difference maps) without degrading the overall visual quality or coherence of the scene.}
\label{fig:supp_case3_generalization}
\end{figure*}

\paragraph{Generalization Across Diverse Scenes.}
While the previous visualizations demonstrated the variety of attack strategies emerging for a single input, it is equally important to verify that our attacker does not overfit to specific image types. In \Cref{fig:supp_case3_generalization}, we present a comprehensive gallery of adversarial examples generated across a wide range of visual domains, including indoor environments, street scenes, and sports activities.

\end{document}